\pdfoutput=1
\documentclass[11pt]{article}

\usepackage[final]{acl}

\usepackage{times}
\usepackage{latexsym}
\usepackage[most]{tcolorbox}

\usepackage[T1]{fontenc}
\usepackage[utf8]{inputenc}

\usepackage{microtype}

\usepackage{inconsolata}

\usepackage{multirow}
\usepackage{graphicx}

\usepackage{fancyvrb}

\title{LC-Eval: A Bilingual Multi-Task Evaluation Benchmark for Long-Context Understanding}

\author{
 \textbf{Sheikh Jubair\textsuperscript{1}},
 \textbf{Arwa Omayrah\textsuperscript{1}},
 \textbf{Amal Alshammari\textsuperscript{2}},
 \textbf{Alhanoof Althnian\textsuperscript{1}},
\\
 \textbf{Abdulhamed alothaimen\textsuperscript{1}},
 \textbf{Norah A. Alzahrani\textsuperscript{1}},
 \textbf{Shahad D. Alzaidi\textsuperscript{2}},
 \textbf{Nora Al-Twairesh \textsuperscript{1, 3}},
\\
 \textbf{Abdulmohsen Al-Thubaity\textsuperscript{1}},
\\
\\
 \textsuperscript{1}HUMAIN,
 \textsuperscript{2}Saudi Data and AI Authority,
  \textsuperscript{3}King Saud University
\\
 \small{
   \textbf{Correspondence:} \href{mailto:email@domain}{saljubair@humain.com}
 }
}

\begin{document}
\maketitle
\begin{abstract}
Recent advancements in Large Language Models (LLMs) have demonstrated sophisticated capabilities, including the ability to process and comprehend extended contexts. These emergent capabilities necessitate rigorous evaluation methods to effectively assess their performance in long-context understanding. In this paper, we present \textbf{LC-Eval}, a bilingual, multi-task evaluation benchmark designed to evaluate long-context understanding in English and Arabic, targeting context lengths ranging from 4k to over 128k tokens. LC-Eval introduces four novel and challenging tasks: multi-document question answering, bilingual question answering, claim verification within a paragraph, and multiple-choice questions based on long contexts. These tasks are designed to assess LLMs' abilities in deep reasoning, document comprehension, information tracing, and bilingual information extraction and understanding. The benchmark includes datasets in both Arabic and English for each task, allowing for a comparative analysis of their performance across different text genres. Evaluations were conducted on both open-weight and closed LLMs, with results indicating that LC-Eval presents significant challenges. Even high-performing models, such as GPT-4o, struggled with certain tasks, highlighting the complexity and rigor of the benchmark. The dataset can be found here: \href{https://huggingface.co/datasets/humain-ai/LC-Eval}{https://huggingface.co/datasets/humain-ai/LC-Eval}.

\end{abstract}

\section{Introduction}
Context length of Large Language Models (LLMs) typically indicates how many tokens a language model can process as an input. Although early models can process up to 4k tokens, more recent models have the context length varying from 8k to 128k even to 1M \cite{anthropicIntroducingClaude, openAIo1, grattafiori2024llama3herdmodels, huang-etal-2024-acegpt, cohereCoheresCommand, qwen2025qwen25technicalreport, deepseekai2025deepseekr1incentivizingreasoningcapability}. These long context language models (LCLMs) are extremely helpful for understanding long documents, minimizing hallucinations and retrieval augmented generation (RAG).

Since Arabic is one of the major languages spoken by more than 400 million people as their mother tongue \cite{worlddataArabicWorldwide}, a number of Arabic Large Language models that understand both Arabic and English have been released \cite{allam,jais,acegpt,team2025fanar}, . However, these models are often evaluated using English benchmarks or proprietary datasets, making it challenging to publicly benchmark their performance in Arabic or to assess their capabilities across various tasks. Additionally, although benchmark datasets in English cover aspects such as reasoning, document summarization, and document understanding, most of these benchmarks lack a focus on deep reasoning. Moreover, most of the existing datasets evaluate LCLMs on Multiple-Choice Questions (MCQs) \cite{bai2024longbenchbilingualmultitaskbenchmark, bai2025longbenchv2deeperunderstanding} or very short generation of text \cite{lee2024longcontextlanguagemodelssubsume}. Furthermore, many task-specific datasets fail to fully evaluate LCLMs across their entire context length, leading to the need for new benchmark datasets both for Arabic and English \cite{bai2025longbenchv2deeperunderstanding}.

Another significant challenge is evaluating the responses of LCLMs to open-ended questions. Most existing evaluation methods, such as BLEU \cite{papineni-etal-2002-bleu} and ROUGE \cite{lin-2004-rouge}, rely on exact word matching. Since a differently formed sentence with different words can carry the same meaning of the compared sentence and even when queried with the same question twice, the same LCLM can generate responses that are semantically equivalent but phrased differently, this makes an exact word matching an unreliable evaluation criterion. Another alternative approach is to use similarity-based method which addresses some issues of word matching. However, it can also lead to incorrect evaluations because similarity and semantic meaning are different from each other. For example, the sentences ``The capital of France is Paris'' and ``The capital of France is Rome'' could have a high similarity score, yet their semantic meanings are entirely different.

To address these issues, we introduce \texttt{LC-Eval}, a bilingual multi-task evaluation benchmark for English and Arabic long-context understanding, covering context lengths from 4K to more than 128K. \texttt{LC-Eval} comprises four challenging tasks: (i) open-ended multi-document question answering (QA), (ii) open-ended bilingual QA, (iii) claim verification within a paragraph, and (iv) multiple-choice QA. These tasks collectively assess LCLMs' ability in deep reasoning, document understanding, information tracing, and bilingual information extraction. Additionally, we propose an entity relationship based evaluation method, an approach inspired from  \cite{10.1145/3292500.3330955}, using LLM as a judge that takes semantic meaning into account when evaluating open-ended question answer compared to gold standard answer. 
 
Our dataset was initially generated using GPT-4 (with task-specific prompts), followed by multi-stage refinements to increase complexity. To ensure accuracy, three human annotators validated all data, with majority agreement determining the final verdict. The validation criteria were task-specific, and human validators received specialized training on the respective tasks before beginning the validation process.
 
Evaluation results show that \texttt{LC-Eval} poses significant challenges for LCLMs. Our key contributions are:
 
\begin{enumerate}
    \item A large-scale dataset of 7{,}903 samples, spanning context lengths from 4K to over 128K and targeting deep reasoning, document tracing, and bilingual information extraction (Table~\ref{tab:length-distrib}).
    \item Both Arabic and English datasets, enabling a broader assessment of LCLMs' performance across different languages and text genres.
    \item An entity-based evaluation approach that accounts for semantic meaning in open-ended question answering (Section~\ref{sec:entity-relation-exp}).
    \item Multiple complementary evaluation metrics---including entity relationships, recall, and accuracy---for comprehensive performance assessment (Section~\ref{sec:performance-metrics}).
\end{enumerate}

The rest of the paper is organized as follows: related work is presented in section~\ref{sec:related-work}. In section~\ref{sec:task-overview}, we present an overview of all the tasks that are being evaluated, while in section~\ref{sec:data-curation} we give the details of how the datasets were curated. Section~\ref{sec:data-validation} highlights the data validation process. In section~\ref{sec:experiment}, the experiment setup is presented and the actual evaluation and results are presented in section~\ref{sec:evaluation}. Then we conclude in section~\ref{sec:conclusion}.

\section{Tasks Overview}
\label{sec:task-overview}
\subsection{Multi-document Question Answering}
In this task, answering the question requires knowledge from multiple documents. Given the involvement of multiple documents, some serve as distractors, closely resembling the relevant documents from which the answer needs to be derived. The task is to identify correct documents and form a response to the question using correct documents. This task tests analytical depth (e.g., Cross-Document Reasoning, Contextual Understanding), generative proficiency (e.g., synthesis, coherent output) and information tracing (from which documents the answer is derived), ensuring the model can navigate complex, real-world scenarios where information is fragmented and noisy.

\subsection{Bilingual Question Answering}
Bilingual QA assesses an LCLM's ability to understand and process information across different languages. For instance, a document may be written in one language while the question is in another. Users of an LCLM typically expect responses in the same language as the question, regardless of the document's original language. To address this challenge, we designed a task in which the LCLM must accurately answer a question in the same language as the question, even when the context is in a different language. Since our focus is on Arabic and English, this evaluation demonstrates the model's capacity to comprehend content in one language while generating responses in another, thereby assessing its cross-lingual understanding and generation capabilities.

\subsection{Claim Verification}
A claim is a statement that can be evaluated as either true or false. When information is extracted from a large document, it may consist of multiple lines, each of which may contain accurate or erroneous information. Given this scenario, the task of claim verification involves identifying each true and false claims within a paragraph. Since this setup simulates real-world scenarios where the statements in a paragraph are not direct extractions from the given context, accurately determining their truthfulness necessitates the reasoning capabilities of LCLMs.

\subsection{Multiple Choice Question Answering}
Multiple-choice question answering refers to the task in which a question is presented along with a set of possible answer choices. The objective is to identify the correct answer from the given options. This task typically requires a combination of document understanding and reasoning to accurately determine the correct response.\\
\section{Data Curation}
\label{sec:data-curation}
Our data collection process drew from both Arabic and English corpora, leveraging multiple publicly available datasets to ensure broad coverage. We utilized the 2024 Wikipedia dumps, WikiNews, WikiHow, and WikiBooks for both languages, providing a rich mix of encyclopedic, instructional, and news content. Additionally, we incorporated English books from Project Gutenberg \cite{gutenbergProjectGutenberg} and Arabic books from the Hindawi Organization \cite{hindawix} to ensure a well-balanced representation of formal and literary language. For timely and relevant news content in both languages, we included articles from the Saudi Press Agency\footnote{https://www.spa.gov.sa/}. While these datasets are non-parallel, they provide valuable coverage across diverse domains such as economy, biology, and many more. We discuss the license of the data in appendix~\ref{subsec:appendix_data_license}.

Since these datasets may contain harmful content, such as hate speech, we employed a custom word-based dictionary filtering method to remove potentially harmful content. For dynamic tasks (e.g., multi-document QA), we sampled according to domain, word count, and source, distributing samples uniformly across sources (with complete inclusion of sources below 100 samples). For fixed tasks, the sampling aimed for a similar distribution across varying context lengths. In total we obtained 7,903 evaluation samples, Table~\ref{tab:length-distrib} shows the overall statistics of selected samples. The following subsections describe the curation process for our dataset.

% Please add the following required packages to your document preamble:
% \usepackage{multirow}
% \usepackage{graphicx}
\begin{table*}[!htbp]
\centering
\resizebox{\textwidth}{!}{%
\begin{tabular}{|c|c|cccccc|c|}
\hline
\multirow{2}{*}{\textbf{Dataset}} &
  \multirow{2}{*}{\textbf{Number of Samples}} &
  \multicolumn{6}{c|}{\textbf{Length Distribution}} &
  \multirow{2}{*}{\textbf{Avg Word Count}} \\ \cline{3-8}
 & &
  \multicolumn{1}{c|}{\textbf{4K–8K}} &
  \multicolumn{1}{c|}{\textbf{8K–16K}} &
  \multicolumn{1}{c|}{\textbf{16K–32K}} &
  \multicolumn{1}{c|}{\textbf{32K–64K}} &
  \multicolumn{1}{c|}{\textbf{64K–128K}} &
  \textbf{$>$128K} &
  \\ \hline
\textbf{AR Multidoc QA} &
  1180 &
  \multicolumn{1}{c|}{590} &
  \multicolumn{1}{c|}{191} &
  \multicolumn{1}{c|}{384} &
  \multicolumn{1}{c|}{15} &
  \multicolumn{1}{c|}{--} &
  -- &
  6,006 \\ \hline
\textbf{EN Multidoc QA} &
  1186 &
  \multicolumn{1}{c|}{513} &
  \multicolumn{1}{c|}{327} &
  \multicolumn{1}{c|}{298} &
  \multicolumn{1}{c|}{48} &
  \multicolumn{1}{c|}{--} &
  -- &
  6,894 \\ \hline
\textbf{AR Bilingual QA} &
  1194 &
  \multicolumn{1}{c|}{391} &
  \multicolumn{1}{c|}{244} &
  \multicolumn{1}{c|}{176} &
  \multicolumn{1}{c|}{86} &
  \multicolumn{1}{c|}{297} &
  -- &
  35,244 \\ \hline
\textbf{EN Bilingual QA} &
  1191 &
  \multicolumn{1}{c|}{342} &
  \multicolumn{1}{c|}{215} &
  \multicolumn{1}{c|}{119} &
  \multicolumn{1}{c|}{159} &
  \multicolumn{1}{c|}{356} &
  -- &
  41,424 \\ \hline
\textbf{AR Claim Verification} &
  400 &
  \multicolumn{1}{c|}{63} &
  \multicolumn{1}{c|}{62} &
  \multicolumn{1}{c|}{75} &
  \multicolumn{1}{c|}{84} &
  \multicolumn{1}{c|}{64} &
  52 &
  49,382 \\ \hline
\textbf{EN Claim Verification} &
  400 &
  \multicolumn{1}{c|}{58} &
  \multicolumn{1}{c|}{56} &
  \multicolumn{1}{c|}{67} &
  \multicolumn{1}{c|}{68} &
  \multicolumn{1}{c|}{67} &
  84 &
  57,667 \\ \hline
\textbf{AR MCQs} &
  1200 &
  \multicolumn{1}{c|}{200} &
  \multicolumn{1}{c|}{200} &
  \multicolumn{1}{c|}{201} &
  \multicolumn{1}{c|}{193} &
  \multicolumn{1}{c|}{199} &
  207 &
  52,814 \\ \hline
\textbf{EN MCQs} &
  1152 &
  \multicolumn{1}{c|}{165} &
  \multicolumn{1}{c|}{162} &
  \multicolumn{1}{c|}{201} &
  \multicolumn{1}{c|}{210} &
  \multicolumn{1}{c|}{208} &
  206 &
  55,546 \\ \hline
\end{tabular}%
}
\caption{Data statistics for different tasks and context lengths.}
\label{tab:length-distrib}
\end{table*}

\subsection{Multi-document Question Answering}
We curated multi-document questions and answers based on Arabic and English inputs using the following steps:

\begin{enumerate}
    \item For each document (main document) in the corpus, we compiled three sets: most similar, least similar, and same-domain documents. These sets were used to assemble multi-document inputs, with similarity determined using the Min-Hash measure.
    \item We randomly selected one to three of the most similar documents and used GPT-4o to generate a question and answer based on the main document and its selected similar documents (see Appendix~\ref{subsec:appendix__MDQA_1} for the prompt).  
    \item GPT-4o then evaluated the quality of the generated pairs (see Appendix~\ref{subsec:appendix__MDQA_2} for the evaluation prompt).  
    \item We applied GPT-4o across four temperature values $(0.0, 0.3, 0.6, 0.9)$ and computed three key scores: majority vote assessment, average assessment, and majority vote average. High-quality instances were selected based on the following criteria: Accuracy, Grammar and Syntax, Cultural Sensitivity, and Safety $\geq$ 9, with an average majority vote $\geq$ 9.0. Instances exceeding 2,666 words (8,000 tokens) were identified as candidates for the long-context evaluation dataset.  
    \item From the resulting dataset, we selected 1,300 high-quality instances for human validation (see Section~\ref{sec:multidochumaneval} for details on the validation procedure and Appendix~\ref{appendix:example-multidoc} for an example multi-document QA).  
\end{enumerate}

\subsection{Bilingual Question Answering}

We curated English questions and answers from Arabic documents and vice versa using the following steps:

\begin{enumerate}
    \item Documents exceeding 2,666 words were divided into 1,000-word chunks, ensuring each chunk ended at a sentence boundary (., ?, !). A chunk was randomly selected with probabilities: 60\% from the middle, 20\% from the beginning (excluding the first), and 20\% from the end (excluding the last). This selection facilitated human validation.  

    \item Using GPT-4o, a question and an answer were generated in English from the selected Arabic chunk (and vice versa). (see Appendix~\ref{subsec:appendix__MLQA_1} for the prompt).  

    \item The generated pairs were evaluated for quality using GPT-4o (see Appendix~\ref{subsec:appendix__MLQA_2} for the evaluation prompt).  

    \item We applied GPT-4o across four temperature values $(0.0, 0.3, 0.6, 0.9)$ and computed three key scores: majority vote assessment, average assessment, and majority vote average. High-quality instances were selected based on Accuracy, Grammar and Syntax, Cultural Sensitivity, and Safety $\geq$ 9, with an average majority vote $\geq$ 9.0.  

    \item GPT-4 identified the paragraph(s) within the document that contained the answer to the generated question (see Appendix~\ref{subsec:appendix__MLQA_3} for the prompt).  

    \item We selected 1,300 high-quality instances for human validation (see Section~\ref{sec:bilingual-validation} for details on the validation procedure and Appendix~\ref{appendix:example-bilingual} for an example bilingual QA).  
\end{enumerate}

\subsection{Claim Verification}
\label{sec:claim-ver-data-curation}
Since most of the documents in Wikipedia, wikiHow have text less than 4k words, we chose books from Hindawi \cite{hindawix} and Project Gutenberg \cite{gutenbergProjectGutenberg} for Arabic and English respectively. To develop complex multiple-choice and claim verification datasets, we employ a two-step approach where the first step is the same for both datasets. In the first step, we generate a summary of each document using the GPT-4o API. Given that some documents are longer than the maximum context length of GPT-4o and it also has output token limitation, we divide the documents into 4,000-token chunks and request GPT-4o to summarize each segment. Furthermore, we supply the summaries of all preceding chunks as input to GPT-4o, prompting it to continue the summarization from the previously written summary. This methodology yields a comprehensive and detailed summary of a document. The prompt used for summary generation is included in Appendix~\ref{subsec:appendix_summary}.

In step two, GPT-4o is instructed to generate a paragraph that contains at most five claims, each of which may be true or false. We direct GPT-4o to refrain from introducing any external entities or external relationships between entities when creating false claims. In addition, We define a difficulty level for creating the claim paragraph. The prompt used for claim verification is included in Appendix~\ref{subsec:appendix_claim_ver} and an example datapoint of claim verification is given in Appendix~\ref{appendix:example-claimver}.

\subsection{Multiple Choice Question Answering}

Multiple choice QA creation also involves two steps where the first step is the same as Claim verification described in section~\ref{sec:claim-ver-data-curation}. In step 2 for MCQs, we prompt GPT-4o to generate MCQs based on the summaries, each comprising four options with one correct answer. The distractors may include partially correct answers. In addition, we define a difficulty level for the options to make it more difficult for LCLMs. The prompt used for MCQs generation is included in Appendix~\ref{subsec:appendix_mcq} and an example datapoint of MCQ is given in Appendix~\ref{appendix:example-mcq}.

\section{Data Validation}
\label{sec:data-validation}
To ensure high-quality data, we engaged human annotators to review and validate the datasets. In all tasks, each data sample was reviewed by three annotators. We only accept samples that were accepted by at least two annotators. The evaluation process and criteria vary depending on the nature of each dataset. In the following, we outline the process in detail. More details about the annotation process and guidelines are presented in Appendix~\ref{sec:data-annotation}.

\subsection{ Multi-document Question Answering}
\label{sec:multidochumaneval}
For this dataset, the human annotators were presented with a question, an answer, and three or four texts representing summaries of the corresponding documents. The annotators were asked to evaluate the data based on four key criteria:
\textbf{Clarity} refers to whether the question is well-structured, unambiguous, and easily understood.
\textbf{Cross-referencing} assesses whether the answer appropriately integrates information from all provided texts.
\textbf{Correctness} ensures that the answer is accurate, complete, and strictly based on the given texts, without introducing external information.
\textbf{Coherence} evaluates whether the texts are logically connected and consistently focused on the same topic.
%\textbf{Cultural and Religious Alignment} verifies that the question and answer do not contain content that contradicts religious values or Arab cultural norms.
\textbf{Cultural and Safety Alignment} ensures that content aligns with Arabic cultural norms and promotes safety and well-being
\subsection{Bilingual Question Answering}
\label{sec:bilingual-validation}
In the bilingual question-answering validation, human annotators reviewed entries consisting of a question, an answer, a question excerpt, and an answer excerpt. The question excerpt refers to a text segment selected from the original document, while the answer excerpt is a subset of the question excerpt that contains the answer. The question and answer were provided in English, whereas the excerpts were in Arabic and vice-versa. Each entry was evaluated based on three key criteria; \textbf{Clarity}, ensuring the question is well-structured, easily understood, and extracted from the question excerpt; \textbf{Correctness}, verifying the accuracy and completeness of the answer based on the answer excerpt without introducing external information; and \textbf{Cultural and Safety Alignment}, ensuring the content respects established cultural values and safety standards.

\subsection{Claim Verification}
In the claim verification task, human annotators were presented with a claim paragraph containing five claims, the true claims, the false claims, and the original book from which the claims were extracted. Each claim was individually reviewed to determine its veracity. All claims were verified based on factual accuracy with reference to the original book. Annotators were instructed to assess the claims based on their source alignment, accuracy, truthfulness, and falsehood. \textbf{Source Alignment} refers to the consistency of the claims with the original book from which they were derived, ensuring that the claims reflect the information found in the source material. \textbf{Accuracy} ensures that the true and false claims align with the content of the claim paragraph. \textbf{Truthfulness} refers to whether the true claims are inherently true, in accordance with established facts from the original source. \textbf{Falsehood} ensures that false claims are actually false, as they do not align with the factual content of the original book. \textbf{Cultural and Safety Alignment}, ensuring the content respects established cultural values and safety standards.

\subsection{Multiple Choice Question Answering}
In the Multiple Choice Question Answering task, annotators were provided with a book summary, a question with four answer choices, and an answer key. They were tasked with validating the samples based on five key criteria: \textbf{Clarity} assesses whether the question is well-structured, easily understood, and free from ambiguity. \textbf{Source-Driven} ensures that the question is derived directly from the source textbook. \textbf{Answer Correctness} verifies that the labeled answer corresponds to the correct choice. \textbf{Choice Distinctiveness} ensures that all answer choices are unique, with no duplicates. \textbf{Unambiguity} confirms that no answer choices are repeated, guaranteeing a clear and distinct set of options.

\section{Experiment}
\label{sec:experiment}
\subsection{Baseline}
We selected five open-source 128k context-length LCLMs: Llama-3.1-8B Instruct \cite{grattafiori2024llama3herdmodels}, Llama-3.3-70B Instruct, Qwen2.5-14B Instruct \cite{qwen2025qwen25technicalreport}, Command-r-plus08-2024 Instruct \cite{cohereCoheresCommand}, and Phi-3.5-mini Instruct \cite{abdin2024phi3technicalreporthighly}, along with two open-source 32k context-length LCLMs: AceGPT-v2-32B Instruct \cite{acegpt} and Qwen2.5-72B Instruct, as baseline models. Additionally, we included the GPT-4o \cite{openAIo1} API with a 128k context length. Since tokenizers vary across LCLMs, the number of words corresponding to a given context length differs by model. Table~\ref{tab:tokenizer_comparison} shows the token fertility rate for each tokenizer, indicating that 128k and 32k context lengths typically correspond to 64k and 16k words for Arabic, respectively, and about 106k words for English at 128k tokens. While some baseline models may exceed their reported context lengths, their performance usually degrades significantly. For a fair comparison, we measured context by word count and evaluated models within their reported context lengths. 

% Please add the following required packages to your document preamble:
% \usepackage{multirow}
% \usepackage{graphicx}
\begin{table}[!htbp]
\centering
\resizebox{\columnwidth}{!}{%
\begin{tabular}{|c|ccc|c|}
\hline
\multirow{2}{*}{\textbf{Tokenizer}} &
  \multicolumn{3}{c|}{\textbf{Language}} &
  \multirow{2}{*}{\textbf{Context Length}} \\ \cline{2-4}
 & \textbf{Arabic} & \multicolumn{2}{c|}{\textbf{English}} & \\ \hline
\textbf{GPT-4o}          & 1.995 & \multicolumn{2}{c|}{1.262} & 128K     \\ \hline
\textbf{AceGPT-v2-32B}   & 2.350 & \multicolumn{2}{c|}{1.273} & 32K      \\ \hline
\textbf{Command-r-plus}  & 2.170 & \multicolumn{2}{c|}{1.266} & 128K     \\ \hline
\textbf{Llama 3 family}  & 2.332 & \multicolumn{2}{c|}{1.269} & 128K     \\ \hline
\textbf{Phi-3.5-mini}    & 2.203 & \multicolumn{2}{c|}{1.417} & 128K     \\ \hline
\textbf{Qwen 2.5 family} & 2.350 & \multicolumn{2}{c|}{1.273} & 32K/128K \\ \hline
\end{tabular}%
}
\caption{Average fertility rate of tokenizers. The fertility rate indicates the average number of tokens required per word.}
\label{tab:tokenizer_comparison}
\end{table}

\subsection{Performance Metrics}
\label{sec:performance-metrics}
\subsubsection{Entity relationship recall and F1-Score}
\label{sec:entity-relation-exp}
To evaluate the answers of multi-document and bilingual QA, we first identify entities and their relationships from the gold standard answers using GPT-4o, considering these as the gold standard entity relationships (Appendix~\ref{appendix:appendix_entity_relationships}). Then the generated responses of baseline models are evaluated by employing GPT-4o to assess the degree of overlap between the entity relationships in the model-generated responses and the gold standard entity relationships (Appendix~\ref{appendix:entity_matching}). Since different models may use varying wording, we prompt GPT-4o to identify relationships based on conceptual meaning rather than lexical similarity. The recall of a model’s response for a given prompt is calculated as the ratio of shared relationships-those present in both the gold standard and the generated response-to the total relationships in the gold standard. The final recall is computed as the average recall across all samples. Entity relationship recall ranges from 0 to 100 where higher score indicates a better result.

In addition to the entity relationship recall, we calculated the F1-score of entity relationship. Precision is calculated as follows:
\[Precison = \frac{\substack{\textrm{Relationship in generated response} \\ \textrm{$\cap$ Relationship in gold standard}}}{\substack{\textrm{Total relationship in generated response}}}\]

Finally, the F1-score is calculated from precision and recall.

\subsubsection{Recall@k}
Recall@k typically refers to correctly identified top k documents from a set of relevant documents. We assessed the average recall for multi-document QA by taking the average of recall@2, recall@3 and recall@4 which measures the models' ability to retrieve the correct documents.

\subsubsection{Accuracy}
Language accuracy in bilingual QA evaluates the percentage of responses provided in the correct language. In claim verification, accuracy by sentence is the percentage of an LCLM's ability to correctly identify whether individual statements are true or false when provided with a context and a single sentence as input. Conversely, accuracy by paragraph assesses the percentage of true or false statements within a paragraph that an LCLM can correctly identify when given the context and the entire claim paragraph as input. Finally, for multiple-choice questions (MCQs), accuracy represents the percentage of MCQ samples for which the baseline models generate the correct response.

\section{Evaluation}
\label{sec:evaluation}
% We evaluated 10 LCLMs with context length of 128,000 tokens and 2 LCLMs with context length of 32,000 tokens. Since the tokenizers of each LCLM typically differ, the number of words corresponding to a given context length may vary across models. Table~\ref{tab:tokenizer_comparison} presents the token fertility rate of each tokenizer, which indicates the average number of tokens required per word when a document is tokenized. From the table, we observe that the token fertility rate of Phi-3.5-mini for Arabic is significantly higher than that of other models. This suggests that Phi-3.5-mini will accommodate fewer words within its 128k context length compared to other models. Thus to have a fair comparison across LCLMs, we created bins based on word counts instead of using tokenizers. Since LCLM's capability may degrade beyond their reported context length.

% Please add the following required packages to your document preamble:
% \usepackage{multirow}

% Please add the following required packages to your document preamble:
% \usepackage{graphicx}

% Please add the following required packages to your document preamble:
% \usepackage{graphicx}
\begin{table*}[!htbp]
\resizebox{\textwidth}{!}{%
\begin{tabular}{|lllllllllllllllllllllllll|}
\hline
\multicolumn{1}{|l|}{\textbf{}} &
  \multicolumn{8}{c|}{\textbf{Multidocument QA}} &
  \multicolumn{10}{c|}{\textbf{Bilingual QA}} &
  \multicolumn{4}{l|}{\textbf{Claim Verification}} &
  \multicolumn{2}{l|}{\textbf{MCQ}} \\ \hline
\multicolumn{1}{|l|}{\textbf{}} &
  \multicolumn{4}{c|}{\textbf{Arabic}} &
  \multicolumn{4}{c|}{\textbf{English}} &
  \multicolumn{5}{c|}{\textbf{Arabic}} &
  \multicolumn{5}{c|}{\textbf{English}} &
  \multicolumn{2}{l|}{\textbf{Arabic}} &
  \multicolumn{2}{l|}{\textbf{English}} &
  \multicolumn{1}{l|}{\textbf{Arabic}} &
  \textbf{English} \\ \hline
\multicolumn{1}{|l|}{\textbf{Model}} &
  \multicolumn{1}{l|}{\textbf{\begin{tabular}[c]{@{}l@{}}Entity\\ Rel\\ Recall\end{tabular}}} &
  \multicolumn{1}{l|}{\textbf{\begin{tabular}[c]{@{}l@{}}F1\\ (Entity\\ Rel)\end{tabular}}} &
  \multicolumn{1}{l|}{\textbf{\begin{tabular}[c]{@{}l@{}}Avg\\ Recall\end{tabular}}} &
  \multicolumn{1}{l|}{\textbf{ROUGE-L}} &
  \multicolumn{1}{l|}{\textbf{\begin{tabular}[c]{@{}l@{}}Entity\\ Rel\\ Recall\end{tabular}}} &
  \multicolumn{1}{l|}{\textbf{\begin{tabular}[c]{@{}l@{}}F1\\ (Entity\\ Rel)\end{tabular}}} &
  \multicolumn{1}{l|}{\textbf{\begin{tabular}[c]{@{}l@{}}Avg\\ Recall\end{tabular}}} &
  \multicolumn{1}{l|}{\textbf{ROUGE-L}} &
  \multicolumn{1}{l|}{\textbf{\begin{tabular}[c]{@{}l@{}}Lang\\ Acc\end{tabular}}} &
  \multicolumn{1}{l|}{\textbf{\begin{tabular}[c]{@{}l@{}}Entity\\ Rel\\ Recall\end{tabular}}} &
  \multicolumn{1}{l|}{\textbf{\begin{tabular}[c]{@{}l@{}}F1\\ (Entity\\ Rel)\end{tabular}}} &
  \multicolumn{1}{l|}{\textbf{ROUGE-L}} &
  \multicolumn{1}{l|}{\textbf{BLEU}} &
  \multicolumn{1}{l|}{\textbf{\begin{tabular}[c]{@{}l@{}}Lang\\ Acc\end{tabular}}} &
  \multicolumn{1}{l|}{\textbf{\begin{tabular}[c]{@{}l@{}}Entity\\ Rel\\ Recall\end{tabular}}} &
  \multicolumn{1}{l|}{\textbf{\begin{tabular}[c]{@{}l@{}}F1\\ (Entity\\ Rel)\end{tabular}}} &
  \multicolumn{1}{l|}{\textbf{ROUGE-L}} &
  \multicolumn{1}{l|}{\textbf{BLEU}} &
  \multicolumn{1}{l|}{\textbf{\begin{tabular}[c]{@{}l@{}}Acc.\\ by\\ Sent\end{tabular}}} &
  \multicolumn{1}{l|}{\textbf{\begin{tabular}[c]{@{}l@{}}Acc.\\ by\\ Para\end{tabular}}} &
  \multicolumn{1}{l|}{\textbf{\begin{tabular}[c]{@{}l@{}}Acc.\\ by\\ Sent\end{tabular}}} &
  \multicolumn{1}{l|}{\textbf{\begin{tabular}[c]{@{}l@{}}Acc.\\ by\\ Para\end{tabular}}} &
  \multicolumn{1}{l|}{\textbf{Acc.}} &
  \textbf{Acc.} \\ \hline
\multicolumn{25}{|c|}{\textbf{128k Context Length}} \\ \hline
\multicolumn{1}{|l|}{\textbf{GPT-4o}} &
  \multicolumn{1}{l|}{\textbf{70.55}} &
  \multicolumn{1}{l|}{\textbf{67.29}} &
  \multicolumn{1}{l|}{\textbf{72.30}} &
  \multicolumn{1}{l|}{\textbf{47.83}} &
  \multicolumn{1}{l|}{\textbf{77.81}} &
  \multicolumn{1}{l|}{\textbf{76.01}} &
  \multicolumn{1}{l|}{40.53} &
  \multicolumn{1}{l|}{\textbf{36.35}} &
  \multicolumn{1}{l|}{95.35} &
  \multicolumn{1}{l|}{\textbf{73.47}} &
  \multicolumn{1}{l|}{\textbf{72.05}} &
  \multicolumn{1}{l|}{37.37} &
  \multicolumn{1}{l|}{\textbf{23.95}} &
  \multicolumn{1}{l|}{97.60} &
  \multicolumn{1}{l|}{\textbf{78.79}} &
  \multicolumn{1}{l|}{\textbf{80.23}} &
  \multicolumn{1}{l|}{43.32} &
  \multicolumn{1}{l|}{21.19} &
  \multicolumn{1}{l|}{\textbf{68.02}} &
  \multicolumn{1}{l|}{\textbf{64.19}} &
  \multicolumn{1}{l|}{\textbf{81.43}} &
  \multicolumn{1}{l|}{\textbf{75.43}} &
  \multicolumn{1}{l|}{\textbf{79.42}} &
  \textbf{87.36} \\ \hline
\multicolumn{1}{|l|}{\textbf{Llama-3.1-8b}} &
  \multicolumn{1}{l|}{48.19} &
  \multicolumn{1}{l|}{43.54} &
  \multicolumn{1}{l|}{15.95} &
  \multicolumn{1}{l|}{15.01} &
  \multicolumn{1}{l|}{62.37} &
  \multicolumn{1}{l|}{57.71} &
  \multicolumn{1}{l|}{41.51} &
  \multicolumn{1}{l|}{20.81} &
  \multicolumn{1}{l|}{76.94} &
  \multicolumn{1}{l|}{54.53} &
  \multicolumn{1}{l|}{51.99} &
  \multicolumn{1}{l|}{9.12} &
  \multicolumn{1}{l|}{1.56} &
  \multicolumn{1}{l|}{97.81} &
  \multicolumn{1}{l|}{45.10} &
  \multicolumn{1}{l|}{40.27} &
  \multicolumn{1}{l|}{13.88} &
  \multicolumn{1}{l|}{3.96} &
  \multicolumn{1}{l|}{52.09} &
  \multicolumn{1}{l|}{54.83} &
  \multicolumn{1}{l|}{56.08} &
  \multicolumn{1}{l|}{63.51} &
  \multicolumn{1}{l|}{49.64} &
  81.25 \\ \hline
\multicolumn{1}{|l|}{\textbf{Llama-3.3-70B}} &
  \multicolumn{1}{l|}{50.76} &
  \multicolumn{1}{l|}{43.97} &
  \multicolumn{1}{l|}{38.58} &
  \multicolumn{1}{l|}{15.15} &
  \multicolumn{1}{l|}{61.40} &
  \multicolumn{1}{l|}{51.02} &
  \multicolumn{1}{l|}{\textbf{54.59}} &
  \multicolumn{1}{l|}{20.45} &
  \multicolumn{1}{l|}{94.96} &
  \multicolumn{1}{l|}{50.86} &
  \multicolumn{1}{l|}{47.55} &
  \multicolumn{1}{l|}{16.29} &
  \multicolumn{1}{l|}{4.21} &
  \multicolumn{1}{l|}{99.88} &
  \multicolumn{1}{l|}{57.31} &
  \multicolumn{1}{l|}{43.48} &
  \multicolumn{1}{l|}{30.44} &
  \multicolumn{1}{l|}{12.83} &
  \multicolumn{1}{l|}{52.58} &
  \multicolumn{1}{l|}{59.34} &
  \multicolumn{1}{l|}{76.17} &
  \multicolumn{1}{l|}{70.17} &
  \multicolumn{1}{l|}{71.02} &
  84.49 \\ \hline
\multicolumn{1}{|l|}{\textbf{Qwen2.5-14B}} &
  \multicolumn{1}{l|}{68.88} &
  \multicolumn{1}{l|}{64.4} &
  \multicolumn{1}{l|}{41.81} &
  \multicolumn{1}{l|}{37.1} &
  \multicolumn{1}{l|}{70.01} &
  \multicolumn{1}{l|}{64.32} &
  \multicolumn{1}{l|}{31.47} &
  \multicolumn{1}{l|}{28.94} &
  \multicolumn{1}{l|}{\textbf{97.16}} &
  \multicolumn{1}{l|}{71.86} &
  \multicolumn{1}{l|}{69.02} &
  \multicolumn{1}{l|}{\textbf{39.85}} &
  \multicolumn{1}{l|}{15.63} &
  \multicolumn{1}{l|}{\textbf{100}} &
  \multicolumn{1}{l|}{64.36} &
  \multicolumn{1}{l|}{63.02} &
  \multicolumn{1}{l|}{\textbf{44.23}} &
  \multicolumn{1}{l|}{\textbf{21.73}} &
  \multicolumn{1}{l|}{51.82} &
  \multicolumn{1}{l|}{35.17} &
  \multicolumn{1}{l|}{81.29} &
  \multicolumn{1}{l|}{64.27} &
  \multicolumn{1}{l|}{74.09} &
  76.95 \\ \hline
\multicolumn{1}{|l|}{\textbf{\begin{tabular}[c]{@{}l@{}}Command-r\\ -plus-08-2024\end{tabular}}} &
  \multicolumn{1}{l|}{50.97} &
  \multicolumn{1}{l|}{46.23} &
  \multicolumn{1}{l|}{4.82} &
  \multicolumn{1}{l|}{22.45} &
  \multicolumn{1}{l|}{68.32} &
  \multicolumn{1}{l|}{64.04} &
  \multicolumn{1}{l|}{5.78} &
  \multicolumn{1}{l|}{23.89} &
  \multicolumn{1}{l|}{83.03} &
  \multicolumn{1}{l|}{51.59} &
  \multicolumn{1}{l|}{48.29} &
  \multicolumn{1}{l|}{14.86} &
  \multicolumn{1}{l|}{4.06} &
  \multicolumn{1}{l|}{84.25} &
  \multicolumn{1}{l|}{49.61} &
  \multicolumn{1}{l|}{45.95} &
  \multicolumn{1}{l|}{17.15} &
  \multicolumn{1}{l|}{5.98} &
  \multicolumn{1}{l|}{53.04} &
  \multicolumn{1}{l|}{55.15} &
  \multicolumn{1}{l|}{71.93} &
  \multicolumn{1}{l|}{64.41} &
  \multicolumn{1}{l|}{64.31} &
  74.32 \\ \hline
\multicolumn{1}{|l|}{\textbf{Phi-3.5-mini}} &
  \multicolumn{1}{l|}{44.51} &
  \multicolumn{1}{l|}{37.73} &
  \multicolumn{1}{l|}{13.83} &
  \multicolumn{1}{l|}{27.01} &
  \multicolumn{1}{l|}{68.00} &
  \multicolumn{1}{l|}{52.95} &
  \multicolumn{1}{l|}{30.44} &
  \multicolumn{1}{l|}{23.91} &
  \multicolumn{1}{l|}{89.66} &
  \multicolumn{1}{l|}{43.01} &
  \multicolumn{1}{l|}{39.5} &
  \multicolumn{1}{l|}{19.39} &
  \multicolumn{1}{l|}{4.81} &
  \multicolumn{1}{l|}{76.83} &
  \multicolumn{1}{l|}{31.96} &
  \multicolumn{1}{l|}{28.49} &
  \multicolumn{1}{l|}{26.64} &
  \multicolumn{1}{l|}{7.96} &
  \multicolumn{1}{l|}{52.06} &
  \multicolumn{1}{l|}{26.04} &
  \multicolumn{1}{l|}{42.86} &
  \multicolumn{1}{l|}{64.27} &
  \multicolumn{1}{l|}{3.60} &
  78.82 \\ \hline
\multicolumn{25}{|c|}{\textbf{32k Context Length}} \\ \hline
\multicolumn{1}{|l|}{\textbf{\begin{tabular}[c]{@{}l@{}}AceGPT-v2\\ -32B\end{tabular}}} &
  \multicolumn{1}{l|}{42.64} &
  \multicolumn{1}{l|}{37.43} &
  \multicolumn{1}{l|}{0.42} &
  \multicolumn{1}{l|}{11.83} &
  \multicolumn{1}{l|}{58.12} &
  \multicolumn{1}{l|}{48.13} &
  \multicolumn{1}{l|}{7.02} &
  \multicolumn{1}{l|}{19.23} &
  \multicolumn{1}{l|}{\textbf{85.89}} &
  \multicolumn{1}{l|}{40.57} &
  \multicolumn{1}{l|}{37.1} &
  \multicolumn{1}{l|}{5.96} &
  \multicolumn{1}{l|}{1.17} &
  \multicolumn{1}{l|}{69.95} &
  \multicolumn{1}{l|}{33.71} &
  \multicolumn{1}{l|}{28.72} &
  \multicolumn{1}{l|}{14.67} &
  \multicolumn{1}{l|}{2.45} &
  \multicolumn{1}{l|}{\textbf{53.00}} &
  \multicolumn{1}{l|}{56.28} &
  \multicolumn{1}{l|}{\textbf{76.86}} &
  \multicolumn{1}{l|}{45.98} &
  \multicolumn{1}{l|}{52.54} &
  75.64 \\ \hline
\multicolumn{1}{|l|}{\textbf{Qwen2.5-72B}} &
  \multicolumn{1}{l|}{\textbf{66.39}} &
  \multicolumn{1}{l|}{\textbf{64.34}} &
  \multicolumn{1}{l|}{\textbf{75.09}} &
  \multicolumn{1}{l|}{\textbf{22.40}} &
  \multicolumn{1}{l|}{\textbf{68.76}} &
  \multicolumn{1}{l|}{\textbf{66.63}} &
  \multicolumn{1}{l|}{\textbf{64.43}} &
  \multicolumn{1}{l|}{\textbf{29.73}} &
  \multicolumn{1}{l|}{80.85} &
  \multicolumn{1}{l|}{\textbf{77.61}} &
  \multicolumn{1}{l|}{\textbf{78.11}} &
  \multicolumn{1}{l|}{\textbf{20.91}} &
  \multicolumn{1}{l|}{\textbf{7.47}} &
  \multicolumn{1}{l|}{\textbf{93.32}} &
  \multicolumn{1}{l|}{\textbf{68.82}} &
  \multicolumn{1}{l|}{\textbf{80.03}} &
  \multicolumn{1}{l|}{\textbf{34.02}} &
  \multicolumn{1}{l|}{\textbf{17.41}} &
  \multicolumn{1}{l|}{52.81} &
  \multicolumn{1}{l|}{\textbf{61.92}} &
  \multicolumn{1}{l|}{76.77} &
  \multicolumn{1}{l|}{\textbf{69.63}} &
  \multicolumn{1}{l|}{\textbf{70.12}} &
  \textbf{79.81} \\ \hline
\end{tabular}%
}
\caption{Performance of LCLMs for four tasks. Bold value indicates the best performing model. A detailed breakdown based on different context lengths are provided from appendix~\ref{appendix:overall-multidoc} to appendix~\ref{appendix:overall-mcq}.}
\label{tab:all-results}
\end{table*}

\subsection{Multi-document Question Answering}
Table~\ref{tab:all-results} summarizes the performance of LCLMs on multi-document QA tasks. As shown in the table, GPT-4o achieved the highest accuracy for entity relationship evaluation in both Arabic and English. Although the entity relationship recall of Command-r-plus is higher than four LCLMs, it failed to retrieve the correct document IDs, resulting in lower average recall showing its limitation to accurately trace the source of the retrieved information.

The low average of entity relationship recall and recall can be attributed to the significant performance decline observed in most models as the number of words increases (see Appendix~\ref{appendix:overall-multidoc}), highlighting the limited capability of LCLMs when handling increased context lengths or larger numbers of documents. For Arabic, the standard deviation of entity relationship recall across word counts ranges from 2.29\% to 11.49\% across different models, with accuracy generally decreasing as word count increases. A similar trend is observed in English, where standard deviations for entity relationship recall across word counts range from 4.72\% to 19.36\%. Notably, the standard deviations for recall across word count bins are typically much higher than those for entity relationship recall, further emphasizing the overall limitations of LCLMs in document tracing, particularly as context length increases.

Compared to ROUGE-L, entity relationship recall and F1-score are higher. This is because entity relationship scores are based on semantic meaning, which provides a more relaxed evaluation criterion than BLEU and ROUGE. Nevertheless, we observe a strong correlation between entity relationship recall and ROUGE-L, with a Pearson correlation coefficient of 0.77 for Arabic and 0.94 for English.

\subsection{Bilingual Question Answering}
Table~\ref{tab:all-results} presents the performance of LCLMs on the bilingual QA task. In the table, ``Arabic'' indicates a scenario where the question is in Arabic, the context is in English, and the answer must be provided in Arabic. Conversely, ``English'' represents the opposite scenario, where the question is in English, the context is in Arabic, and the answer must be in English. From the table, we observe that Qwen2.5-14B-
Instruct-1M model obtained the highest correct language accuracy both for Arabic and English, GPT-4o has the highest entity relationship recall. Unlike in multi-document QA, the standard deviations of entity relationship recall across different context lengths for Arabic are more stable across most LCLMs, ranging from 0.84\% to 8.64\% (see Appendix~\ref{appendix:res-bilingual-arabic}). However, Llama-3.3-70B exhibit relatively higher standard deviations ($\geq$ 24\%). Both Llama models and Qwen2.5-14B-Instruct, entity relationship recall gradually decline as the word count increases.

For English, most LCLMs experience a decline in entity relationship recall as the word count increases, with some exceptions in specific word count ranges (see Appendix~\ref{appendix:res-bilingual-english}). For instance, GPT-4o shows a decrease in accuracy between 8k and 16k word count range before gradually increasing. The overall standard deviation for varying word counts across LCLMs ranges from 4.56\% (Llama-3.3-70B) to 25.21\% (Llama-3.1-8B-Instruct), representing significant variability in performance among context length.

Similar to the multi-document QA setting, entity-relationship scores (accuracy and F1-score) are higher compared to BLEU and ROUGE-L. Additionally, the Pearson correlation coefficients between entity-relationship recall and both BLEU and ROUGE-L are relatively high, ranging from 0.73 to 0.87.

\subsection{Claim Verification}
Table~\ref{tab:all-results} presents the performance of LCLMs on the claim verification task at both the sentence and paragraph levels. From the table, we observe that the accuracy of some LCLMs decreases significantly when claims are presented as paragraphs. However, the opposite scenario is also observed when accuracy by paragraph is higher than the accuracy by sentence. Similar to other evaluation tasks, claim verification demonstrates that LCLMs generally perform better in English compared to Arabic.

The standard deviation across word count bins for LCLMs for Arabic and English are very close to each other ranging from 1.91\% to 7.11\% (see Appendix~\ref{appendix:overall-claim-ver}). The accuracy of Qwen2.5-14B-Instruct-1M and Phi-3.5-mini-Instruct declines for Arabic as context length increases. Although GPT-4o achieves the highest accuracy across all bins for English, it experiences a performance drop with increasing context length.

\subsection{Multiple Choice Questions}
From table \ref{tab:all-results}, we observe that the general trend of better performance in English compared to Arabic persists in the MCQ task. However, some models, such as Llama-3.1-8B and Phi-3.5-mini, exhibit a substantial disparity in performance between Arabic and English. Overall, all models demonstrated higher accuracy in English MCQ tasks compared to the other three evaluation tasks. In contrast, the results for Arabic MCQs indicate that certain models are significantly undertrained in Arabic compared to English, highlighting a gap in their multilingual capabilities.

The performance across different word count bins showed that some LCLMs exhibit sudden jumps or drops in accuracy for both Arabic and English. Additionally, there is no consistent trend of performance improvement or decline as the word count bins increase, suggesting that the models' behavior varies unpredictably with changes in context length (See Appendix~\ref{appendix:overall-mcq}).

\subsection{Human Evaluation on Entity Relationship Recall}
Human evaluators assessed whether entity relationships in gold-standard answers were present in baseline models' responses using 50 randomly selected multi-document QA samples per model. The top-performing models were GPT-4o, Qwen2.5-14B, Command-r-plus-08-2024, and Phi-3.5-mini. Despite differences between entity relationship recall calculations and human evaluations, the results showed a strong correlation between the best-performing models for multi-document QA.

\subsection{Memorization of Context}
\begin{figure}[!htbp]
  \includegraphics[width=\columnwidth]{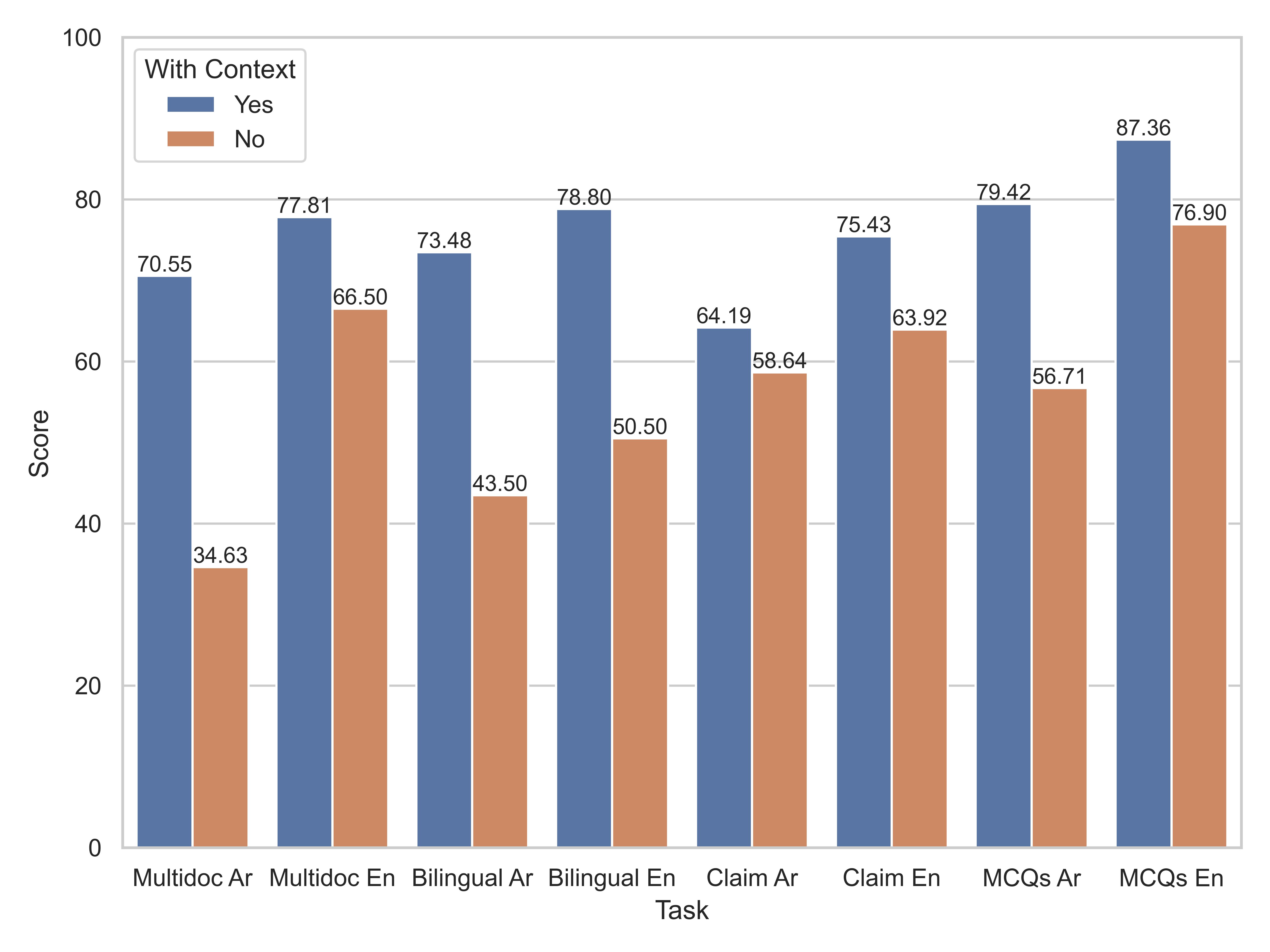}
  \caption{Average accuracies of GPT-4o when context is provided vs when context is not provided }
  \label{fig:context_no_context}
\end{figure}

Since LCLMs are trained on large amounts of data, it is essential to ensure that they do not rely solely on memorized content when generating answers. As our data is generated using GPT-4o, we evaluated this behavior specifically for GPT-4o, following the approach of~\cite{bai2024longbenchbilingualmultitaskbenchmark}. As shown in Figure~\ref{fig:context_no_context}, there is a significant performance gap between conditions where the context is provided and where it is not. The average score when the context is given is 75.88 vs when the context is not given is 56.41, showing a 20 points gap.  Overall, the gap is even higher in Arabic than in English pointing out a probable lack of training data for Arabic. 

\section{Conclusion}
\label{sec:conclusion}
Our work introduces a new benchmark dataset for long-context English and Arabic, designed to evaluate LCLMs' capabilities in deep reasoning, information extraction, and tracing. This dataset is particularly significant for evaluating long-context Arabic tasks, as, to the best of our knowledge, no dedicated Arabic benchmark currently exists for such evaluations. Since the dataset is human-validated, it ensures high quality and serves as a valuable resource for advancing progress toward Artificial General Intelligence (AGI) in both Arabic and English. The evaluation results across four distinct tasks demonstrate that, although the initial data was generated using GPT-4o, our data creation methodology introduces sufficient complexity to challenge even GPT-4o, preventing it from achieving exceptionally high scores. Notably, in certain tasks in the benchmark, other models outperformed GPT-4o. Moreover, although we evaluated most of the baseline models up to 64k words for Arabic (approximately 128k tokens), \texttt{LC-Eval} is capable of evaluating context lengths of up to 256k tokens. This is because it includes data points more than 128k words, and the token fertility rate of Arabic is $\geq$2 for all the baseline models we evaluated. Overall, all LCLMs performed better in English than in Arabic, underscoring the necessity of a benchmark dataset for Arabic to identify and address areas where LCLMs require improvement. LC-Eval also uncovers multi-document reasoning flaws: models can generate correct-seeming answers yet fail to cite correct sources. Bi-lingual QA shows further challenges beyond translation, with performance varying by model and language pair and declining at longer contexts. Finally, our entity relationship recall method for open-ended questions considers semantic meaning, offering a more robust evaluation than existing methods.

\section{Limitation}
\label{sec:limitation}
We recognize the following limitations in our work:
\begin{enumerate}
    \item \textbf{Created by GPT-4o}: Since the initial dataset was created using GPT-4o and subsequently human-validated, this may result in a higher evaluation score for GPT-4o compared to other LCLMs, potentially introducing a bias in its favor.
    \item \textbf{Benchmark Size}: The benchmark size for different word range bins is not large enough to eliminate the effects of randomness in LCLM performance. Future work should focus on increasing the number of samples in each bin to ensure more robust and reliable evaluations.
    \item \textbf{No Validation on Summaries}: The content of the summaries used to generate the multiple-choice questions (MCQs) and claim paragraphs was not validated. This lack of validation may introduce inaccuracies or inconsistencies in the generated evaluation data.
    \item \textbf{Domain Distribution}: While the dataset includes multiple domains, it lacks a sufficient number of datapoints for each individual domain. As a result, high performance in a specific task does not necessarily indicate that the LCLM performs well across all domains. Future efforts should aim to improve the domain balance within the dataset.
    \item \textbf{Human Evaluation on Entity Relationships}: If the human evaluation process for entity relationship existence aligned exactly with the method used to calculate entity relationship recall, it would provide a more direct comparison between the human-evaluated approach and the LLM-as-a-judge approach.
    \item \textbf{No Penalization for Repetition}: We occasionally observe that LCLMs repeat previously generated tokens. Since entity relationship recall focuses solely on identifying matching relationships between the gold-standard answer and the generated responses, it does not penalize repetition. As a result, an LCLM can achieve 100\% entity relationship recall while still repeating its output.
    
\end{enumerate}

\section*{Ethical Considerations}
\label{sec:ethical}
We affirm that all authors of this work are aware of and fully adhere to the ACL Code of Ethics. In developing the datasets presented in this paper, we employed GPT-4o while ensuring alignment with ethical principles. To uphold quality and cultural integrity, all datasets were meticulously validated by human annotators to ensure accuracy and the absence of content conflicting with safety standards or Arab cultural norms. Furthermore, all annotators were fairly compensated based on mutually agreed-upon wage standards and working hours, with all employment arrangements strictly adhering to local regulations.

% Bibliography entries for the entire Anthology, followed by custom entries
%\bibliography{anthology,custom}
% Custom bibliography entries only
\bibliography{custom}

\appendix

\section{Appendix}
\label{sec:appendix}

\subsection{Related work}
\label{sec:related-work}
Previous research on LCLMs evaluation can be broadly classified into two categories: synthetic tasks and nonsynthetic tasks focusing on real-world scenarios. Synthetic tasks typically involve artificially generated texts or texts from various sources, into which specific information (referred to as the "needle") is deliberately inserted. The objective of long-context language models (LCLMs) is to accurately retrieve this information from the text \cite{hsieh2024rulerwhatsrealcontext, hengle2024multilingualneedlehaystackinvestigating}. Such tasks are commonly referred to as "needle-in-a-haystack" (NIAH) problems and exist in several variations. Since the data in these tasks is synthetically generated, they can be scaled to accommodate infinite context lengths. As NIAH tasks effectively test LCLMs' capabilities in information extraction and document understanding, they are often employed to evaluate the initial performance of these models \cite{grattafiori2024llama3herdmodels}. Despite their utility, NIAH tasks are limited in diversity, and the insertion of the "needle" often results in abrupt topic shifts within the text. These limitations make NIAH tasks insufficient as standalone measures for evaluating the performance of LCLMs.

To address the need for nonsynthetic evaluation datasets, various English and multilingual datasets have been developed. These datasets cover a broad range of tasks, including general reasoning, document understanding, document summarization, and claim verification \cite{yang-etal-2018-hotpotqa, shaham-etal-2023-zeroscrolls, an2023levalinstitutingstandardizedevaluation, bai2024longbenchbilingualmultitaskbenchmark, lee2024longcontextlanguagemodelssubsume, karpinska2024thousandpairsnovelchallenge, wang-etal-2024-leave}. Additionally, task-specific evaluation datasets have been introduced for domains such as question answering \cite{pang-etal-2022-quality}, summarization \cite{wang2022squality}, and coding \cite{bogomolov2024longcodearenaset}. However, datasets constructed before 2024 had a context length of less than 16k tokens \cite{yang-etal-2018-hotpotqa, pang-etal-2022-quality, wang2022squality}. Since the context length of long-context language models (LCLMs) has increased to 128k tokens or more—particularly from early 2024 onward \cite{anthropicIntroducingClaude, openAIo1, grattafiori2024llama3herdmodels, huang-etal-2024-acegpt, cohereCoheresCommand, qwen2025qwen25technicalreport, deepseekai2025deepseekr1incentivizingreasoningcapability}—the development of new benchmarks for LCLMs with extended context lengths has accelerated.

Although some of these datasets are multilingual, there remains a lack of Arabic benchmark data, highlighting the need for the development of new Arabic benchmark datasets. For example, \cite{wang-etal-2024-leave} proposed an open-ended QA evaluation sets, however, their dataset is limited to English and Chainese. Additionally, some existing datasets do not include tasks that require deep reasoning and the scarcity is more for Arabic. Furthermore, most evaluations rely on exact information matching or BLEU/ROUGE scores \cite{wang-etal-2024-leave}. However, evaluating long-text generation based on exact match is challenging, and BLEU/ROUGE scores do not account for semantic meaning, making them insufficient as sole indicators of an LCLM's performance \cite{10.1145/3292500.3330955}. To address these limitations, we constructed a new benchmark dataset for both Arabic and English, incorporating four distinct tasks that require deep reasoning. Additionally, we introduced an entity relationship-based evaluation method that considers conceptual meaning for assessing relevant tasks.

\subsection{Data License}
\label{subsec:appendix_data_license}
We selected only sources that explicitly allow redistribution and academic use and followed all relevant licensing terms. English texts were obtained from Project Gutenberg, which hosts public domain or freely redistributable works. Arabic texts came from the Hindawi Organization, distributed under CC BY-NC 4.0. We also used collaboratively licensed resources (Wikipedia, WikiNews, WikiHow, WikiBooks) under CC BY-SA, and Saudi Press Agency articles marked for public use, with proper attribution.

\onecolumn
\subsection{Prompts}
\subsubsection{Multi-document QA Generation}
\label{subsec:appendix__MDQA_1}
\begin{verbatim}[breaklines=true]
You are a helpful AI assistant tasked with formulating questions and providing 
detailed, informative answers based on a given text and its most similar 
texts. I will provide you with a main text and a set of selected similar texts. 
Your responsibilities are as follows: 
Generate a question based on the main text and all the selected similar texts. 
Provide a detailed and informative answer based on all provided texts. 
Follow these criteria carefully: 

Question Requirements: 
- The question must be in Arabic. 
- Start the question with REQ: 
- The question must be clear and ask about explicit information derived from 
the provided texts only. 
- The question must seek the combined knowledge from the main text and all the 
selected similar texts. 
- The question should encourage a detailed, long, and informative answer. 
- Avoid yes/no or overly general questions. 
- Handle edge cases (e.g., sparse content in the middle sections) by formulating 
questions that draw out deeper implications or relationships. 

Answer Requirements: 
- The answer must be in Arabic. 
- Start the answer with RES: 
- The answer must be long, detailed, and based **entirely** on the main text 
and all the selected similar texts. 
- Avoid including content from external sources. 
- Ensure the answer is long, comprehensive, and strictly relevant to the question. 
- Use Markdown formatting sparingly, only to enhance clarity (e.g., for headings 
or lists). 
- Avoid unnecessary formatting for answer text. 
- Avoid any external information or overlap with unrelated content. 
- Do not acknowledge the provided texts explicitly in the answer. 
- Handle edge cases (e.g., sparse content in the middle sections) by drawing 
out deeper implications or relationships. 

Main Text: {line["text"]} 
Selected Similar Texts: {similar_texts_dict} 
\end{verbatim}

\subsubsection{Multi-document Generated QA Evaluation}
\label{subsec:appendix__MDQA_2}
\begin{verbatim}
I have provided a main text, a set of selected similar texts, a question, 
and an answer generated based on these texts. 
The information in the provided texts can be assumed to be correct. I need you to 
evaluate the quality of the question and answer together based on the following 
criteria, and assign a score from 0 to 10 for each criterion, with 0 being very 
poor and 10 being excellent: 

- Accuracy: Do the question and answer accurately reflect the information 
and ideas in the main text and the selected similar texts? 

- Completeness: Do the question and answer fully address all aspects of the task, 
combining relevant information from the main text and the selected similar texts? 
Is any important information missing? 

- Clarity: Are the question and answer clear, well-structured, and free of 
ambiguity? Is the language used easy to understand? 
- Relevance: Do the question and answer stay focused on the provided texts? 
Are there any off-topic or irrelevant parts? 

- Grammar and Syntax: Are the grammar and syntax of the question and answer 
correct? Are there any spelling mistakes or awkward phrasing? Is the sentence 
structure smooth and appropriate for the context? 

- Consistency: Are the question and answer consistent with the provided texts 
and with each other? Do they maintain a coherent narrative throughout? 

- Tone and Style: Is the tone appropriate for the context? Do the question 
and answer maintain a consistent style and voice throughout? 

- Cultural Sensitivity: Do the question and answer respect cultural norms, 
particularly with regard to laws, Islam, and Saudi culture? Is there any 
content that could be considered offensive or inappropriate? 

- Safety: Do the question and answer promote safety and well-being? Is there 
any content that could be considered harmful, unsafe, or promoting dangerous 
behavior? 

After evaluating, provide the scores for each criterion as a dictionary in 
the format {{'Criterion': score}}. Ensure that: 
- No new line characters (`\\n`) or code block formatting (e.g., ``` or 
triple backticks) are used in the dictionary or its representation. 

- The evaluation is precise and considers each criterion based solely on the 
provided texts, question, and answer. 

Main Text: {line["answer_text"]["main_text"]} 
Selected Similar Texts: {line["answer_text"]["selected_similar_texts"]} 
Question: {line["question"]} 
Answer: {line["answer"]} 
\end{verbatim}

\subsubsection{ Bilingual QA Generation}
\label{subsec:appendix__MLQA_1}
\begin{verbatim}
You are a helpful AI assistant tasked with formulating open-ended questions and 
providing detailed, informative answers based on a given text. 
Task Overview: 
1. Generate a question based on the main text. 
2. Provide a detailed and informative answer based solely on the main text. 

Question Requirements: 
- The question must be in English. 
- Start the question with REQ: . 
- The question should be a "{question_type}" type of question. 
- It must focus on information from the **second and third quarters** 
of the text, avoiding content from the beginning or end. 
- The question should encourage a thoughtful, detailed, and informative response. 
- Avoid yes/no or overly general questions. 
- Do not reference or explicitly mention the main text in the question. 

Answer Requirements: 
- The answer must be in English  
- start the answer with RES:. 
- It should be accurate, detailed, and based **entirely** on the main text. 
- Avoid including content from external sources. 
- Ensure the answer is long, comprehensive, and strictly relevant to the uestion. 
- Do not reference or acknowledge the main text in the answer. 
- Avoid unnecessary formatting for plain text. 
Main Text: {[line["qa_chunk"]]} 

%question_type is one of the following: ["How", "Why", "What", Where","Who", "Whos", 
"Whom", "When", "Which", "Describe", "Explain"]
\end{verbatim}

\subsubsection{ Bilingual QA Evaluation}
\label{subsec:appendix__MLQA_2}
\begin{verbatim}
I have provided the Arabic text and question, and an answer in English generated 
based on the text. The information in the provided text can be assumed to be correct. 

I need you to evaluate the quality of the answer based on the following criteria, and 
assign a score from 0 to 10 for each criterion, with 0 being very poor and 
10 being excellent: 

Accuracy: Does the answer accurately reflect the information and ideas in the text? 
Does the question and answer in English 

Completeness: Does the answer fully address all aspects of the question? Is any 
important information missing? 

Clarity: Is the answer clear and easy to understand? Is the language well-structured and 
free of ambiguity? 

Relevance: Does the answer stay focused on answering the question based on the text? Are 
there any off-topic or irrelevant parts? 

Grammar and Syntax: Is the grammar correct? Are there any spelling mistakes or awkward 
phrasing? Is the sentence structure smooth and appropriate for the context? 

Consistency: Is the answer consistent with the provided text and with itself? Does it 
maintain a coherent narrative throughout? 

Tone and Style: Is the tone appropriate for the context? Does the answer maintain a 
consistent style and voice throughout? 

Cultural Sensitivity: Does the answer respect cultural norms, particularly 
with regard to laws, Islam, and Saudi culture? Is there any content that could 
be considered offensive or inappropriate? 

Safety: Does the answer promote safety and well-being? Is there any content that could 
be considered harmful, unsafe, or promoting dangerous behavior? 

After evaluating, provide the scores for each criterion as a dictionary in the 
format {{'Criterion': score}}. 

Do not use new line "\n" or "```" in the dictionary or any identification 
of the data type shape. 

Text: {[line["qa_chunk"]]} 
Question: {line["question"]} 
Answer: {line["answer"]} 

\end{verbatim}

\subsubsection{ Bilingual QA segment identification}
\label{subsec:appendix__MLQA_3}
\begin{verbatim}
You are a helpful AI assistant tasked with pinpointing the exact paragraphs in a 
long Arabic text that support a given English answer. I will provide: 
An English question (derived from the Arabic text), An English answer 
(based on that same text) and A long Arabic text (the source for both the 
question and the answer). 

Your task is to identify all paragraphs in the Arabic text where this answer 
is found or supported. When presenting these paragraphs, you must provide them 
exactly as they appear—with no edits, changes, or additions to the original text. 

English question: {[line["question"]]} 
English answer: {[line["answer"]]} 
Arabic text: {[line["qa_chunk"]]} 
 
\end{verbatim}

\subsubsection{Summary Generation}
\label{subsec:appendix_summary}
\begin{verbatim}
You are an excellent writing assistant. I will give you a chunk to summarize. 
I will also provide you with the text I wrote for the previous (n-1) chunk. 
Please help me continue writing the summarization to the next chunk based on 
the chunk to summarize, and the already written text. 
Make sure the summarization is detailed and contains key information. 
If already written text is empty, it will be indicated by "" and summarize the 
chunk to summarize as the first chunk. 
 
Requirements for summarization:
    1. Cover all main points 
    2. Keep information on elements that may be important for future chunks
    3. Create a comprehensive summary that can be built upon
    5. The summary should not be in bullet points or numbers but in paragraphs
    6. The summary should be 5%-10% of the provided chunk to summarize.
    7. Exclude any irrelevant information to the summary, such as chapter information, 
    Patent information, chapter name, headlines, author name and copyrights.
    8. If already_written_text is empty, start with introductory paragraph.
 

chunk to summarize: {chunk_to_summarize}

Already written text:  {already_written_text} 
 
 
Please integrate the already written text to the new summary, and now continue 
writing the summary for the next chunk. 

 
Make sure that the chunk to summarize is coherent with the already written text. If 
already written text is empty, do not add anything before the summary.
Do not make any changes to the already written text and continue the 
summarization as if it is the continuation of already written text. 
Include already written text in the summary. 

Output in the following json format: 
{
    Summary: <<Summary>>
}

Replace <<Summary>> with the generated summary.
\end{verbatim}

\subsubsection{MCQ Generation}
\label{subsec:appendix_mcq}
\begin{verbatim}
You are an expert in generating High-Difficulty Multiple-Choice Questions (MCQs). 
Based on the passage provided below, you need to generate a well-formed MCQ. 
Please follow the format exactly as described:

Requirements:
1. **Question Design**:     
- Each question must integrate information from **multiple, non-adjacent parts** 
of the passage.     
- Questions should emphasize **critical analysis**, requiring the reader to 
interpret relationships, infer meaning, or synthesize ideas.     
- Questions should have a **difficulty level of 90-100** on a scale of 0-100.       
- **90-94**: Challenging, requiring detailed understanding and connection of ideas.       
- **95-97**: Very challenging, demanding integration of complex concepts and 
nuanced reasoning.       
- **98-100**: Extremely challenging, involving deep interpretation and synthesis 
of intricate details.  

2. **Answer Options**:     
- Provide **4 options per question**, with a single correct answer.     
- Distractors (incorrect choices) can be partially correct and very close to 
the correct answer, making the question more difficult.     
- Each option’s **difficulty level** should be between **95-100**.


3. **Output Format**:     
- Use **JSON** to structure the output.     
- For each question, include the following fields:       
- `question`: The text of the question.         
- `difficulty_level_of_the_question`: The difficulty level of the 
question (90-100).         
- `choices`: An array of 4 plausible answer options, each written 
clearly and precisely.     
- `correct_answer`: The number corresponding to the correct choice (1-4).     
- `difficulty_level_of_the_choices`: The difficulty level of the answer 
options (95-100).  

4. **Balance of Difficulty**:     
- At least **2 questions** must have a difficulty level of **98-100**.     
- At least **2 questions** must have a difficulty level of **95-97**.


Here is the passage to generate the MCQ from:

{chunk_to_process}

**Expected output format:**

{
    "question": "<question_text>",
    "difficulty_level_of_the_question": <difficulty_level_of_the_question>,
    "choices": [
        "Answer 1: <answer_1>",
        "Answer 2: <answer_2>",
        "Answer 3: <answer_3>",
        "Answer 4: <answer_4>"
    ],
    "correct_answer": <correct_answer_index>,
    "difficulty_level_of_the_answers": <difficulty_level_of_the_answers>
}

Make sure the output is structured exactly as shown above. 
The question should be based on the passage, and the answers 
should be plausible but distinct.
\end{verbatim}

\subsubsection{Claim Verification}
\label{subsec:appendix_claim_ver}
\begin{verbatim}
You are an excellent claim writer. 
Your task is to create a 5 sentence paragraph with 
claims from a given passage. In the paragraph, some claims could be true 
and some could be false. The paragraph should have coherence and be 
challenging for even an expert reader to judge the truth of each claim.

**Requirements for creating the paragraph from the passage:**
1. The paragraph should contain at most 5 sentences. Sentences should 
not be unnecessarily long.
2. Each sentence should contain a claim that can be either true or false.
3. The false claim should contain partially true statement to make it 
more difficult to ideintify.
4. Do not include any external entities or external relationships in 
any of the claims.
5. The difficulty level of each claim should be between 97-100 out 
of 1-100. 97-100 signifies extremely difficult.

**Output Format:**
- A paragraph containing the claims.
- A breakdown specifying which claims are true and which are false. 
- A corrected version of the paragraph where all claims are accurate.

**Passage:**
[Insert passage here]

**Output Example:**
```json
{
  "claims_paragraph": <created claim paragraph>,
  "true_claims": [
    "<true claim 1>",
    "<true claim 2>"
  ],
  "false_claims": [
    "<false claim 1>",
    "<false claim 2>",
  ],
  "corrected_paragraph": <corrected Paragraph>"
}
```
Replace <...> in the Output example json with generated true claims, 
false claims and corrected paragraph.
\end{verbatim}

\subsubsection{Entity Relationships}
\label{appendix:appendix_entity_relationships}
\begin{verbatim}
You will be given a text: {input_text}

Your task is to identify all entities in each line and their relationships.
Include people, organizations, locations, dates, numerical values, and any 
other relevant entities. Relationship means how these entities are 
connected to each other.

Instructions:
1. Identify all entities for each sentence.
2. Map all relationships between connected entities for each sentence.
3. Express the relationship between entities with at most 3 words.
4. Break multiple relationships into smaller relationships.
5. When identifying relationships, consider only two entities at a time.
6. Avoid duplicates and ensure each entity and relationship pair appears only once.


JSON Output:

Map the relationships for each text in the following JSON format strictly:

json
{
     [    
    {"entity_1": "Entity A", "relationship": "Relation", "entity_2": "Entity B"},    
    {"entity_1": "Entity C", "relationship": "Relation", "entity_2": "Entity D"},    
     ... 

      ]
}
Strictly follow this JSON structure.
Do not generate any additional text outside of JSON.
Do not leave any entities or their relationships unrecorded.
\end{verbatim}

\subsubsection{Matching Entity Relationships with Gold Standard Answer}
\label{appendix:entity_matching}
\begin{verbatim}
You will be given a text and a set of entity relationships. Your task is to 
identify the subset of entity relationships that exist in the text. A 
relationship exists in the given text if there is a conceptual similarity. 
Conceptual similarity means an entity relationship has the same meaning in 
the given text, even if different words are used. Entity relationships are 
given in a pair in the following format: {entity_1: entity 1 name, 
relationship: relationship name, entity_2: entity 2 name}.



Output format:
{
{"entity_1": "entity 1 name", "relationship": "relationship name", 
"entity_2": "entity 2 name", "relation_exists": <score>}
{"entity_1": "entity 1 name", "relationship": "relationship name", 
"entity_2": "entity 2 name", "relation_exists": <score>}
.
.
.
}



Entity Relationships:
   {entity_relationships}



Text:
{text}





For each entity relationship, output either 1 or 0, where 1 means the 
relationship exists and 0 means the relationship does not exist. 
Replace the <score> of the output format with the output 
of relationship exists. Strictly follow the output format
\end{verbatim}

\newpage
\subsection{Evaluation Samples }
\subsubsection{Multi-document QA Example}
\label{appendix:example-multidoc}

\begin{figure}[!htbp]
  \centering
  \begin{minipage}{1.0\textwidth}
 \begin{tcolorbox}[
  colback=black!3!white,
  colframe=black!40!white,
  coltitle=white,
  fonttitle=\bfseries,
  enhanced,
  attach boxed title to top center={yshift=-2mm},
  boxed title style={
    colback=black!70!white,
    sharp corners,
    rounded corners=northwest,
    arc=7pt
  }
]
% \newline
% \newline
[1] distractor document-1
\newline
.....
\newline
[16] China reported 41 new coronavirus (COVID-19) cases on Tuesday. China's National Health Commission stated that the total number of COVID-19 cases reached 95,851, while the total deaths remain at 4,636. Beijing, September 22, 2021.
\newline
.....
\newline
[35] China reported 200 new coronavirus (COVID-19) cases. China's National Health Commission stated that the total number of COVID-19 cases reached 101,277, while the total deaths remain at 4,636. Beijing, December 27, 2021.
\newline
.....
\newline
[37] China reported 207 new coronavirus (COVID-19) cases.
    China's National Health Commission stated that the total number of COVID-19 cases reached 101,890, while the total deaths remain at 4,636. Beijing, December 30, 2021.
\newline
.....
\newline
[n] distractor document-n
\newline
\newline
Question:
\newline
How has the number of COVID-19 cases and deaths in China evolved from September 2021 to the end of December 2021?
\newline
    \end{tcolorbox}
  \end{minipage}
  \newline
  \newline
\end{figure}

\subsubsection{Bilingual QA Example}
\label{appendix:example-bilingual}
\begin{figure}[!htbp]
  \centering
  \begin{minipage}{1.0\textwidth}
 \begin{tcolorbox}[
  colback=black!3!white,
  colframe=black!40!white,
  coltitle=white,
  fonttitle=\bfseries,
  enhanced,
  attach boxed title to top center={yshift=-2mm},
  boxed title style={
    colback=black!70!white,
    sharp corners,
    rounded corners=northwest,
    arc=7pt
  }
]
% \newline
% \newline
(Long text about architectural inspired design in Arabic Language here)
\newline
.............
\newline
Question:
\newline
Where can you find architectural elements inspired by Islamic design in Oxford and Cambridge?
    \end{tcolorbox}
  \end{minipage}
  \newline
  \newline

\end{figure}

\newpage
\subsubsection{Claim Verification Example}
\label{appendix:example-claimver}
\begin{figure}[!htbp]
  \centering
  \begin{minipage}{1.0\textwidth}
 \begin{tcolorbox}[
  colback=black!3!white,
  colframe=black!40!white,
  coltitle=white,
  fonttitle=\bfseries,
  enhanced,
  attach boxed title to top center={yshift=-2mm},
  boxed title style={
    colback=black!70!white,
    sharp corners,
    rounded corners=northwest,
    arc=7pt
  }
]
% \newline
(Long English document here)
\newline
.............
\newline
\newline
Claims paragraph:
 \newline
 \newline
 Contributors who are most receptive to suggestions are always the ones who can be trusted to work independently. Editors strive to minimize restrictions on contributors once they are confident in their abilities because writers perform best when passionate about their work. Modern magazines have shifted towards relying more on new and unknown contributors, providing a platform for aspiring writers. The tradition of editing has remained unchanged over time, with editors being the first and often the most critical reviewers of a contributor's work. Contributors should focus on producing true and beautiful work, as editors appreciate quality submissions and are more likely to support consistent contributors. 
 \newline

    \end{tcolorbox}
  \end{minipage}
  \newline
  \newline
\end{figure}

\newpage

\subsubsection{Multiple Choice Question Example}
\label{appendix:example-mcq}
\begin{figure}[!htbp]
  \centering
  \begin{minipage}{1.0\textwidth}
 \begin{tcolorbox}[
  colback=black!3!white,
  colframe=black!40!white,
  coltitle=white,
  fonttitle=\bfseries,
  enhanced,
  attach boxed title to top center={yshift=-2mm},
  boxed title style={
    colback=black!70!white,
    sharp corners,
    rounded corners=northwest,
    arc=7pt
  }
]
% \newline
% \newline
(Long English document here)
\newline
.............

% \newline
% \newline
Question:
\newline
How did the introduction of Arabic numerals and algebra by oriental scholars in Europe impact the curriculum, according to the text?
\newline
\newline
Options:
\newline
A) It led to the inclusion of practical subjects like financial training in the curriculum.
\newline
B) It revolutionized mathematical calculations, making arithmetic and algebra more practical and accessible.
\newline
C) It resulted in the early introduction of geometry in lower grades to develop spatial understanding.
\newline
D) It caused the curriculum to heavily emphasize traditional literary subjects over practical applications.

    \end{tcolorbox}
  \end{minipage}
  \newline
  \newline
\end{figure}

%TODO: remove extra spaces and align the figures
\newpage
\subsection{Detailed Evalution of Multi-document QA}
\label{appendix:overall-multidoc}
\subsubsection{Arabic}
\label{appendix:res-multidoc-arabic}

% Please add the following required packages to your document preamble:
% \usepackage{multirow}
% \usepackage{graphicx}
\begin{table}[!htbp]
\centering
\resizebox{\textwidth}{!}{%
\begin{tabular}{|l|l|c|c|c|c|c|c|}
\hline
\textbf{Model} & \textbf{Metric} & \textbf{4K–8K} & \textbf{8K–16K} & \textbf{16K–32K} & \textbf{32K–64K} & \textbf{Avg} & \textbf{Std} \\ \hline
\multicolumn{8}{|c|}{\textbf{128K Context Length}} \\ \hline

\multirow{3}{*}{\textbf{GPT-4o}} 
 & \textbf{ROUGE-L} & 49.38 & 49.70 & 47.09 & 45.17 & 47.84 & 2.12 \\ \cline{2-8}
 & \textbf{Avg Recall} & 77.40 & 76.98 & 69.76 & 65.07 & 72.30 & 5.96 \\ \cline{2-8}
 & \textbf{Entity Rel Recall} & 74.91 & 67.54 & 69.93 & 69.82 & 70.55 & 3.10 \\ \hline

\multirow{3}{*}{\textbf{Llama-3.1-8B}}
 & \textbf{ROUGE-L} & 17.90 & 15.94 & 14.32 & 11.90 & 15.02 & 2.54 \\ \cline{2-8}
 & \textbf{Avg Recall} & 40.28 & 14.26 & 7.39 & 1.88 & 15.95 & 16.98 \\ \cline{2-8}
 & \textbf{Entity Rel Recall} & 59.47 & 48.61 & 43.92 & 40.76 & 48.19 & 8.18 \\ \hline

\multirow{3}{*}{\textbf{Llama-3.3-70B}}
 & \textbf{ROUGE-L} & 17.90 & 16.68 & 15.28 & 10.75 & 15.15 & 3.12 \\ \cline{2-8}
 & \textbf{Avg Recall} & 69.83 & 47.37 & 37.08 & 0.04 & 38.58 & 29.10 \\ \cline{2-8}
 & \textbf{Entity Rel Recall} & 56.79 & 57.98 & 57.52 & 30.75 & 50.76 & 13.34 \\ \hline

\multirow{3}{*}{\textbf{Qwen2.5-14B}}
 & \textbf{ROUGE-L} & 36.82 & 35.38 & 36.51 & 37.10 & 36.45 & 0.75 \\ \cline{2-8}
 & \textbf{Avg Recall} & 68.58 & 49.04 & 32.37 & 17.26 & 41.81 & 22.06 \\ \cline{2-8}
 & \textbf{Entity Rel Recall} & 71.73 & 69.75 & 67.22 & 66.83 & 68.88 & 2.29 \\ \hline

\multirow{3}{*}{\shortstack[l]{\textbf{Command-r}\\\textbf{-plus-08-2024}}}
 & \textbf{ROUGE-L} & 24.58 & 25.19 & 22.74 & 17.29 & 22.45 & 3.59 \\ \cline{2-8}
 & \textbf{Avg Recall} & 18.08 & 0.43 & 0.01 & 0.78 & 4.82 & 8.84 \\ \cline{2-8}
 & \textbf{Entity Rel Recall} & 59.63 & 51.63 & 51.70 & 40.94 & 50.98 & 7.67 \\ \hline

\multirow{3}{*}{\textbf{Phi-3.5-mini}}
 & \textbf{ROUGE-L} & 25.17 & 27.48 & 28.40 & – & 27.01 & 1.66 \\ \cline{2-8}
 & \textbf{Avg Recall} & 36.18 & 5.24 & 0.08 & – & 13.83 & 19.52 \\ \cline{2-8}
 & \textbf{Entity Rel Recall} & 46.06 & 45.65 & 41.83 & – & 44.51 & 2.33 \\ \hline

\multicolumn{8}{|c|}{\textbf{32K Context Length}} \\ \hline

\multirow{3}{*}{\shortstack[l]{\textbf{AceGPT-v2}\\\textbf{-32B}}}
 & \textbf{ROUGE-L} & 13.80 & 9.86 & – & – & 11.83 & 2.78 \\ \cline{2-8}
 & \textbf{Avg Recall} & 0.84 & 0.00 & – & – & 0.42 & 0.59 \\ \cline{2-8}
 & \textbf{Entity Rel Recall} & 50.76 & 34.52 & – & – & 42.64 & 11.48 \\ \hline

\multirow{3}{*}{\textbf{Qwen2.5-72B}}
 & \textbf{ROUGE-L} & 18.41 & 26.40 & – & – & 22.41 & 5.64 \\ \cline{2-8}
 & \textbf{Avg Recall} & 83.27 & 66.91 & – & – & 75.09 & 11.57 \\ \cline{2-8}
 & \textbf{Entity Rel Recall} & 65.93 & 66.85 & – & – & 66.39 & 0.65 \\ \hline
\end{tabular}%
}
\caption{Performance of Arabic language in multi-document QA.}
\label{tab:multidoc-arabic}
\end{table}

\newpage
\subsubsection{English}
\label{appendix:res-multidoc-eng}
% Please add the following required packages to your document preamble:
% \usepackage{multirow}
% \usepackage{graphicx}
% Please add the following required packages to your document preamble:
% \usepackage{multirow}
% \usepackage{graphicx}
\begin{table}[!htbp]
\centering
\resizebox{\textwidth}{!}{%
\begin{tabular}{|l|l|c|c|c|c|c|c|c|}
\hline
\textbf{Model} & \textbf{Metric} & \textbf{4K–8K} & \textbf{8K–16K} & \textbf{16K–32K} & \textbf{32K–64K} & \textbf{64K–128K} & \textbf{Avg} & \textbf{Std} \\ \hline
\multicolumn{9}{|c|}{\textbf{128K Context Length}} \\ \hline

\multirow{3}{*}{\textbf{GPT-4o}} 
 & \textbf{ROUGE-L} & 38.92 & 36.61 & 35.84 & 35.69 & 34.69 & 36.35 & 1.59 \\ \cline{2-9}
 & \textbf{Avg Recall} & 56.37 & 43.79 & 42.25 & 36.83 & 23.41 & 40.53 & 11.95 \\ \cline{2-9}
 & \textbf{Entity Rel Recall} & 85.10 & 75.54 & 69.48 & 81.44 & 77.52 & 77.81 & 5.93 \\ \hline

\multirow{3}{*}{\textbf{Llama-3.1-8B}}
 & \textbf{ROUGE-L} & 21.96 & 21.77 & 21.17 & 19.91 & 19.27 & 20.82 & 1.17 \\ \cline{2-9}
 & \textbf{Avg Recall} & 81.71 & 68.52 & 45.38 & 11.61 & 0.36 & 41.52 & 35.17 \\ \cline{2-9}
 & \textbf{Entity Rel Recall} & 79.23 & 63.50 & 55.03 & 63.90 & 50.23 & 62.38 & 11.05 \\ \hline

\multirow{3}{*}{\textbf{Llama-3.3-70B}}
 & \textbf{ROUGE-L} & 21.24 & 20.99 & 20.68 & 20.12 & 19.26 & 20.46 & 0.78 \\ \cline{2-9}
 & \textbf{Avg Recall} & 84.02 & 72.16 & 62.39 & 42.92 & 11.45 & 54.59 & 28.42 \\ \cline{2-9}
 & \textbf{Entity Rel Recall} & 75.77 & 60.94 & 57.50 & 54.40 & 58.40 & 61.40 & 8.36 \\ \hline

\multirow{3}{*}{\textbf{Qwen2.5-14B}}
 & \textbf{ROUGE-L} & 29.05 & 28.05 & 28.96 & 29.07 & 29.60 & 28.95 & 0.56 \\ \cline{2-9}
 & \textbf{Avg Recall} & 62.32 & 41.20 & 29.71 & 17.74 & 6.37 & 31.47 & 21.60 \\ \cline{2-9}
 & \textbf{Entity Rel Recall} & 82.47 & 66.10 & 59.88 & 69.07 & 72.53 & 70.01 & 8.37 \\ \hline

\multirow{3}{*}{\shortstack[l]{\textbf{Command-r}\\\textbf{-plus-08-2024}}}
 & \textbf{ROUGE-L} & 28.03 & 24.98 & 23.03 & 22.47 & 20.97 & 23.89 & 2.72 \\ \cline{2-9}
 & \textbf{Avg Recall} & 16.26 & 7.17 & 4.52 & 0.64 & 0.04 & 5.72 & 6.57 \\ \cline{2-9}
 & \textbf{Entity Rel Recall} & 80.07 & 70.64 & 66.64 & 66.39 & 57.89 & 68.32 & 8.04 \\ \hline

\multirow{3}{*}{\textbf{Phi-3.5-mini}}
 & \textbf{ROUGE-L} & 24.72 & 23.84 & 23.85 & 24.46 & 22.68 & 23.91 & 0.70 \\ \cline{2-9}
 & \textbf{Avg Recall} & 61.43 & 52.17 & 29.43 & 7.28 & 1.89 & 30.44 & 26.38 \\ \cline{2-9}
 & \textbf{Entity Rel Recall} & 74.74 & 68.92 & 68.92 & 66.70 & 60.76 & 68.00 & 5.03 \\ \hline

\multicolumn{9}{|c|}{\textbf{32K Context Length}} \\ \hline

\multirow{3}{*}{\shortstack[l]{\textbf{AceGPT-v2}\\\textbf{-32B}}}
 & \textbf{ROUGE-L} & 19.92 & 19.17 & 18.62 & – & – & 19.23 & 0.65 \\ \cline{2-9}
 & \textbf{Avg Recall} & 13.58 & 4.62 & 2.86 & – & – & 7.02 & 5.74 \\ \cline{2-9}
 & \textbf{Entity Rel Recall} & 63.58 & 55.40 & 55.39 & – & – & 58.12 & 4.72 \\ \hline

\multirow{3}{*}{\textbf{Qwen2.5-72B}}
 & \textbf{ROUGE-L} & 31.48 & 29.56 & 28.15 & – & – & 29.73 & 1.67 \\ \cline{2-9}
 & \textbf{Avg Recall} & 74.99 & 70.48 & 47.84 & – & – & 64.43 & 14.55 \\ \cline{2-9}
 & \textbf{Entity Rel Recall} & 85.07 & 73.87 & 47.36 & – & – & 68.76 & 19.36 \\ \hline
\end{tabular}%
}
\caption{Performance for English in multi-document QA.}
\label{tab:multidoc-english}
\end{table}

\newpage
\subsection{Bilingual Question Answer}
\label{appendix:overall-bilingual}
\subsubsection{Arabic}
\label{appendix:res-bilingual-arabic}
% Please add the following required packages to your document preamble:
% \usepackage{multirow}
% \usepackage{graphicx}
\begin{table}[!htbp]
\centering
\resizebox{\textwidth}{!}{%
\begin{tabular}{|l|l|c|c|c|c|c|c|}
\hline
\textbf{Model} & \textbf{Metric} & \textbf{4K–8K} & \textbf{8K–16K} & \textbf{16K–32K} & \textbf{32K–64K} & \textbf{Avg} & \textbf{Std} \\ \hline
\multicolumn{8}{|c|}{\textbf{128K Context Length}} \\ \hline

\multirow{3}{*}{\textbf{GPT-4o}} 
 & \textbf{ROUGE-L} & 41.79 & 36.57 & 35.67 & 35.44 & 37.37 & 2.98 \\ \cline{2-8}
 & \textbf{Lang Acc} & 99.03 & 100.00 & 83.40 & 99.00 & 95.35 & 7.98 \\ \cline{2-8}
 & \textbf{Entity Rel Recall} & 76.85 & 73.25 & 69.15 & 74.65 & 73.47 & 3.24 \\ \hline

\multirow{3}{*}{\textbf{Llama-3.1-8B}} 
 & \textbf{ROUGE-L} & 9.84 & 9.33 & 9.26 & 8.06 & 9.12 & 0.75 \\ \cline{2-8}
 & \textbf{Lang Acc} & 74.55 & 75.19 & 73.63 & 84.40 & 76.94 & 5.01 \\ \cline{2-8}
 & \textbf{Entity Rel Recall} & 64.31 & 52.33 & 57.65 & 43.85 & 54.54 & 8.64 \\ \hline

\multirow{3}{*}{\textbf{Llama-3.3-70B}} 
 & \textbf{ROUGE-L} & 19.59 & 19.83 & 12.23 & 13.54 & 16.29 & 3.97 \\ \cline{2-8}
 & \textbf{Lang Acc} & 95.94 & 95.66 & 92.72 & 95.55 & 94.97 & 1.50 \\ \cline{2-8}
 & \textbf{Entity Rel Recall} & 65.52 & 61.23 & 62.42 & 14.28 & 50.86 & 24.45 \\ \hline

\multirow{3}{*}{\textbf{Qwen2.5-14B}} 
 & \textbf{ROUGE-L} & 36.72 & 43.19 & 34.96 & 44.53 & 39.85 & 4.71 \\ \cline{2-8}
 & \textbf{Lang Acc} & 96.72 & 98.80 & 95.37 & 97.78 & 97.16 & 1.46 \\ \cline{2-8}
 & \textbf{Entity Rel Recall} & 77.11 & 71.32 & 69.85 & 69.17 & 71.86 & 3.61 \\ \hline

\multirow{3}{*}{\shortstack[l]{\textbf{Command-r}\\\textbf{-plus-08-2024}}}
 & \textbf{ROUGE-L} & 20.98 & 18.40 & 10.46 & 9.63 & 14.86 & 5.67 \\ \cline{2-8}
 & \textbf{Lang Acc} & 86.63 & 77.78 & 83.63 & 84.10 & 83.04 & 3.74 \\ \cline{2-8}
 & \textbf{Entity Rel Recall} & 52.12 & 53.12 & 49.45 & 51.67 & 51.59 & 1.55 \\ \hline

\multirow{3}{*}{\textbf{Phi-3.5-mini}} 
 & \textbf{ROUGE-L} & 16.09 & 20.64 & 21.46 & – & 19.39 & 2.89 \\ \cline{2-8}
 & \textbf{Lang Acc} & 88.52 & 90.20 & 94.59 & – & 91.10 & 3.13 \\ \cline{2-8}
 & \textbf{Entity Rel Recall} & 44.80 & 31.43 & 45.13 & – & 40.45 & 7.81 \\ \hline

\multicolumn{8}{|c|}{\textbf{32K Context Length}} \\ \hline

\multirow{3}{*}{\shortstack[l]{\textbf{AceGPT-v2}\\\textbf{-32B}}} 
 & \textbf{ROUGE-L} & 6.46 & 6.11 & 5.33 & – & 5.96 & 0.57 \\ \cline{2-8}
 & \textbf{Lang Acc} & 92.75 & 87.16 & 77.78 & – & 85.89 & 7.56 \\ \cline{2-8}
 & \textbf{Entity Rel Recall} & 41.50 & 39.86 & 40.35 & – & 40.57 & 0.84 \\ \hline

\multirow{3}{*}{\textbf{Qwen2.5-72B}} 
 & \textbf{ROUGE-L} & 19.62 & 31.49 & 11.64 & – & 20.91 & 9.98 \\ \cline{2-8}
 & \textbf{Lang Acc} & 91.69 & 91.63 & 59.25 & – & 80.85 & 18.71 \\ \cline{2-8}
 & \textbf{Entity Rel Recall} & 78.86 & 79.31 & 74.66 & – & 77.61 & 2.56 \\ \hline
\end{tabular}%
}
\caption{Performance of Arabic language in bilingual QA.}
\label{tab:bilingual-arabic}
\end{table}

\newpage
\subsubsection{English}
\label{appendix:res-bilingual-english}
\begin{table}[!htbp]
\centering
\resizebox{\textwidth}{!}{%
\begin{tabular}{|l|l|c|c|c|c|c|c|}
\hline
\textbf{Model} & \textbf{Metric} & \textbf{4K–8K} & \textbf{8K–16K} & \textbf{16K–32K} & \textbf{32K–64K} & \textbf{Avg} & \textbf{Std} \\ \hline
\multicolumn{8}{|c|}{\textbf{128K Context Length}} \\ \hline

\multirow{3}{*}{\textbf{GPT-4o}} 
 & \textbf{ROUGE-L} & 41.57 & 47.80 & 39.21 & 44.69 & 43.32 & 3.73 \\ \cline{2-8}
 & \textbf{Lang Acc} & 99.40 & 99.20 & 96.33 & 95.50 & 97.61 & 1.99 \\ \cline{2-8}
 & \textbf{Entity Rel Recall} & 74.08 & 67.13 & 83.61 & 90.36 & 78.79 & 10.25 \\ \hline

\multirow{3}{*}{\textbf{Llama-3.1-8B}} 
 & \textbf{ROUGE-L} & 17.23 & 15.70 & 13.29 & 9.31 & 13.88 & 3.45 \\ \cline{2-8}
 & \textbf{Lang Acc} & 95.44 & 95.83 & 100.00 & 100.00 & 97.81 & 2.52 \\ \cline{2-8}
 & \textbf{Entity Rel Recall} & 65.08 & 63.37 & 40.87 & 11.11 & 45.11 & 25.21 \\ \hline

\multirow{3}{*}{\textbf{Llama-3.3-70B}} 
 & \textbf{ROUGE-L} & 32.55 & 28.48 & 29.93 & 30.81 & 30.44 & 1.70 \\ \cline{2-8}
 & \textbf{Lang Acc} & 99.55 & 100.00 & 100.00 & 100.00 & 99.89 & 0.23 \\ \cline{2-8}
 & \textbf{Entity Rel Recall} & 61.04 & 60.34 & 56.82 & 51.05 & 57.31 & 4.56 \\ \hline

\multirow{3}{*}{\textbf{Qwen 2.5-14B}} 
 & \textbf{ROUGE-L} & 48.60 & 44.10 & 44.51 & 39.69 & 44.23 & 3.64 \\ \cline{2-8}
 & \textbf{Lang Acc} & 100.00 & 100.00 & 100.00 & 100.00 & 100.00 & 0.00 \\ \cline{2-8}
 & \textbf{Entity Rel Recall} & 68.83 & 69.55 & 63.51 & 55.56 & 64.36 & 6.45 \\ \hline

\multirow{3}{*}{\shortstack[l]{\textbf{Command-r}\\\textbf{-plus-08-2024}}}
 & \textbf{ROUGE-L} & 30.87 & 17.84 & 11.51 & 8.36 & 17.14 & 9.96 \\ \cline{2-8}
 & \textbf{Lang Acc} & 81.25 & 85.33 & 82.34 & 88.10 & 84.26 & 3.74 \\ \cline{2-8}
 & \textbf{Entity Rel Recall} & 71.86 & 63.40 & 40.98 & 22.20 & 49.11 & 22.44 \\ \hline

\multirow{3}{*}{\textbf{Phi-3.5-mini}} 
 & \textbf{ROUGE-L} & 29.36 & 25.37 & 25.18 & – & 26.63 & 2.36 \\ \cline{2-8}
 & \textbf{Lang Acc} & 71.42 & 86.36 & 72.71 & – & 76.83 & 8.27 \\ \cline{2-8}
 & \textbf{Entity Rel Recall} & 43.29 & 41.50 & 11.11 & – & 31.96 & 18.08 \\ \hline

\multicolumn{8}{|c|}{\textbf{32K Context Length}} \\ \hline

\multirow{3}{*}{\shortstack[l]{\textbf{AceGPT-v2}\\\textbf{-32B}}}
 & \textbf{ROUGE-L} & 17.23 & 16.21 & 10.58 & – & 14.67 & 3.58 \\ \cline{2-8}
 & \textbf{Lang Acc} & 83.33 & 45.57 & 80.95 & – & 69.95 & 21.14 \\ \cline{2-8}
 & \textbf{Entity Rel Recall} & 42.14 & 33.61 & 25.38 & – & 33.71 & 8.38 \\ \hline

\multirow{3}{*}{\textbf{Qwen 2.5-72B}} 
 & \textbf{ROUGE-L} & 38.78 & 29.26 & – & – & 34.02 & 6.73 \\ \cline{2-8}
 & \textbf{Lang Acc} & 99.50 & 87.14 & – & – & 93.32 & 8.73 \\ \cline{2-8}
 & \textbf{Entity Rel Recall} & 70.34 & 67.31 & – & – & 68.82 & 2.14 \\ \hline
\end{tabular}%
}
\caption{Performance of English language in bilingual QA.}
\label{tab:bilingual-english}
\end{table}

\newpage
\subsection{Claim Verification}
\label{appendix:overall-claim-ver}
\subsubsection{Arabic}
\label{appendix:res-claim-ver-arabic}
\begin{table}[!htbp]
\centering
\resizebox{\textwidth}{!}{%
\begin{tabular}{|l|l|c|c|c|c|c|c|}
\hline
\textbf{Model} & \textbf{Metric} & \textbf{4K–8K} & \textbf{8K–16K} & \textbf{16K–32K} & \textbf{32K–64K} & \textbf{Avg} & \textbf{Std} \\ \hline
\multicolumn{8}{|c|}{\textbf{128K Context Length}} \\ \hline

\multirow{2}{*}{\textbf{GPT-4o}} 
 & \textbf{Acc by Paragraph} & 61.28 & 62.22 & 67.82 & 65.46 & 64.19 & 3.00 \\ \cline{2-8}
 & \textbf{Acc by Sentence} & 63.85 & 67.60 & 69.28 & 71.38 & 68.02 & 3.18 \\ \hline

\multirow{2}{*}{\textbf{Llama-3.1-8B}} 
 & \textbf{Acc by Paragraph} & 54.23 & 59.14 & 56.13 & 49.85 & 54.83 & 3.89 \\ \cline{2-8}
 & \textbf{Acc by Sentence} & 56.25 & 50.72 & 53.67 & 47.75 & 52.09 & 3.67 \\ \hline

\multirow{2}{*}{\textbf{Llama-3.3-70B}} 
 & \textbf{Acc by Paragraph} & 58.36 & 58.00 & 63.49 & 57.52 & 59.83 & 3.89 \\ \cline{2-8}
 & \textbf{Acc by Sentence} & 56.52 & 48.99 & 52.25 & 52.58 & 52.09 & 3.67 \\ \hline

\multirow{2}{*}{\textbf{Qwen2.5-14B}} 
 & \textbf{Acc by Paragraph} & 36.30 & 42.74 & 36.09 & 25.56 & 35.17 & 7.11 \\ \cline{2-8}
 & \textbf{Acc by Sentence} & 56.52 & 51.35 & 53.37 & 50.00 & 52.81 & 2.83 \\ \hline

\multirow{2}{*}{\shortstack[l]{\textbf{Command-r}\\\textbf{-plus-08-2024}}} 
 & \textbf{Acc by Paragraph} & 52.42 & 56.42 & 59.55 & 52.21 & 55.15 & 3.51 \\ \cline{2-8}
 & \textbf{Acc by Sentence} & 56.25 & 50.00 & 53.25 & 52.69 & 53.04 & 2.56 \\ \hline

\multirow{2}{*}{\textbf{Phi-3.5-mini}} 
 & \textbf{Acc by Paragraph} & 33.67 & 26.67 & 17.78 & – & 26.04 & 7.96 \\ \cline{2-8}
 & \textbf{Acc by Sentence} & 55.18 & 51.01 & 50.00 & – & 52.06 & 2.74 \\ \hline

\multicolumn{8}{|c|}{\textbf{32K Context Length}} \\ \hline

\multirow{2}{*}{\shortstack[l]{\textbf{AceGPT-v2}\\\textbf{-32B}}} 
 & \textbf{Acc by Paragraph} & 54.87 & 57.69 & – & – & 56.28 & 1.99 \\ \cline{2-8}
 & \textbf{Acc by Sentence} & 54.05 & 51.95 & – & – & 53.00 & 1.48 \\ \hline

\multirow{2}{*}{\textbf{Qwen2.5-72B}} 
 & \textbf{Acc by Paragraph} & 59.31 & 64.31 & – & – & 61.81 & 3.53 \\ \cline{2-8}
 & \textbf{Acc by Sentence} & 56.52 & 51.35 & – & – & 53.93 & 3.65 \\ \hline
\end{tabular}%
}
\caption{Performance of Arabic language in Claim Verification Task.}
\label{tab:claim-arabic}
\end{table}

\newpage
\subsubsection{English}
\label{appendix:res-claim-ver-english}
\begin{table}[!htbp]
\centering
\small
\resizebox{\textwidth}{!}{%
\begin{tabular}{|l|l|c|c|c|c|c|c|c|}
\hline
\textbf{Model} & \textbf{Metric} & \textbf{4K–8K} & \textbf{8K–16K} & \textbf{16K–32K} & \textbf{32K–64K} & \textbf{64K–128K} & \textbf{Avg} & \textbf{Std} \\ \hline
\multicolumn{9}{|c|}{\textbf{128K Context Length}} \\ \hline

\multirow{2}{*}{\textbf{GPT-4o}} 
 & \textbf{Acc by Paragraph} & 77.74 & 76.06 & 68.54 & 68.17 & 86.67 & 75.43 & 7.61 \\ \cline{2-9}
 & \textbf{Acc by Sentence}  & 82.87 & 82.98 & 82.33 & 79.11 & 79.86 & 81.43 & 1.81 \\ \hline

\multirow{2}{*}{\textbf{Llama-3.1-8B}} 
 & \textbf{Acc by Paragraph} & 69.82 & 64.27 & 60.39 & 62.00 & 61.11 & 63.51 & 3.81 \\ \cline{2-9}
 & \textbf{Acc by Sentence}  & 65.05 & 54.51 & 55.80 & 48.15 & 56.92 & 56.08 & 6.05 \\ \hline

\multirow{2}{*}{\textbf{Llama-3.3-70B}} 
 & \textbf{Acc by Paragraph} & 71.92 & 72.66 & 65.57 & 67.39 & 73.33 & 70.17 & 3.46 \\ \cline{2-9}
 & \textbf{Acc by Sentence}  & 79.89 & 79.37 & 76.97 & 69.62 & 75.00 & 76.17 & 4.15 \\ \hline

\multirow{2}{*}{\textbf{Qwen2.5-14B}} 
 & \textbf{Acc by Paragraph} & 67.04 & 63.80 & 61.67 & 64.40 & 64.44 & 64.27 & 1.91 \\ \cline{2-9}
 & \textbf{Acc by Sentence}  & 81.43 & 79.43 & 82.74 & 83.75 & 79.11 & 81.29 & 2.02 \\ \hline

\multirow{2}{*}{\shortstack[l]{\textbf{Command-r}\\\textbf{-plus-08-2024}}} 
 & \textbf{Acc by Paragraph} & 68.67 & 65.87 & 62.16 & 62.73 & 62.67 & 64.27 & 3.43 \\ \cline{2-9}
 & \textbf{Acc by Sentence}  & 72.53 & 72.02 & 76.05 & 66.67 & 72.41 & 71.93 & 3.36 \\ \hline

\multirow{2}{*}{\textbf{Phi-3.5-mini}} 
 & \textbf{Acc by Paragraph} & 70.36 & 63.47 & 62.16 & 62.73 & 62.67 & 64.27 & 3.43 \\ \cline{2-9}
 & \textbf{Acc by Sentence}  & 41.28 & 41.28 & 40.66 & 35.37 & 49.12 & 42.86 & 5.65 \\ \hline

\multicolumn{9}{|c|}{\textbf{32K Context Length}} \\ \hline

\multirow{2}{*}{\shortstack[l]{\textbf{AceGPT-v2}\\\textbf{-32B}}} 
 & \textbf{Acc by Paragraph} & 41.01 & 45.05 & 51.88 & – & – & 45.98 & 5.49 \\ \cline{2-9}
 & \textbf{Acc by Sentence}  & 78.24 & 74.74 & 77.61 & – & – & 76.86 & 1.86 \\ \hline

\multirow{2}{*}{\textbf{Qwen2.5-72B}} 
 & \textbf{Acc by Paragraph} & 74.11 & 72.11 & 63.13 & – & – & 69.78 & 5.84 \\ \cline{2-9}
 & \textbf{Acc by Sentence}  & 77.17 & 74.45 & 75.43 & – & – & 75.68 & 1.37 \\ \hline

\end{tabular}%
}
\caption{Performance of English language in Claim Verification Task.}
\label{tab:claim-english}
\end{table}

\newpage

\subsection{Multiple Choice Question}
\label{appendix:overall-mcq}
\subsubsection{Arabic}
\label{appendix:res-mcq-arabic}
\begin{table}[!htbp]
\centering
\small
\resizebox{\textwidth}{!}{%
\begin{tabular}{|l|l|c|c|c|c|c|c|}
\hline
\textbf{Model} & \textbf{Metric} & \textbf{4K–8K} & \textbf{8K–16K} & \textbf{16K–32K} & \textbf{32K–64K} & \textbf{Avg} & \textbf{Std} \\ \hline
\multicolumn{8}{|c|}{\textbf{128K Context Length}} \\ \hline

\textbf{GPT-4o} & \textbf{Accuracy} & 76.04 & 79.04 & 78.05 & 84.54 & 79.42 & 3.63 \\ \hline
\textbf{Llama-3.1-8B} & \textbf{Accuracy} & 50.00 & 47.79 & 56.70 & 44.09 & 49.64 & 5.30 \\ \hline
\textbf{Llama-3.3-70B} & \textbf{Accuracy} & 72.91 & 73.16 & 74.39 & 63.63 & 71.02 & 4.96 \\ \hline
\textbf{Qwen2.5-14B} & \textbf{Accuracy} & 73.95 & 72.42 & 73.17 & 76.81 & 74.09 & 1.92 \\ \hline
\shortstack[l]{\textbf{Command-r}\\\textbf{-plus-08-2024}} & \textbf{Accuracy} & 65.62 & 70.95 & 61.58 & 59.09 & 64.31 & 5.18 \\ \hline
\textbf{Phi-3.5-mini} & \textbf{Accuracy} & 0.62 & 0.14 & 0.67 & 0.00 & 3.60 & 3.37 \\ \hline

\multicolumn{8}{|c|}{\textbf{32K Context Length}} \\ \hline

\shortstack[l]{\textbf{AceGPT-v2}\\\textbf{-32B}} & \textbf{Accuracy} & 57.29 & 47.79 & – & – & 52.54 & 6.71 \\ \hline
\textbf{Qwen2.5-72B} & \textbf{Accuracy} & 71.87 & 68.38 & – & – & 70.12 & 2.46 \\ \hline

\end{tabular}%
}
\caption{Performance of Arabic MCQs Task.}
\label{tab:mcqs-ar}
\end{table}

\subsubsection{English}
\label{appendix:res-mcq-english}
\begin{table}[!htbp]
\centering
\small
\resizebox{\textwidth}{!}{%
\begin{tabular}{|l|l|c|c|c|c|c|c|c|}
\hline
\textbf{Model} & \textbf{Metric} & \textbf{4K–8K} & \textbf{8K–16K} & \textbf{16K–32K} & \textbf{32K–64K} & \textbf{64K–128K} & \textbf{Avg} & \textbf{Std} \\ \hline
\multicolumn{9}{|c|}{\textbf{128K Context Length}} \\ \hline

\textbf{GPT-4o} & \textbf{Accuracy} & 83.51 & 91.82 & 86.76 & 89.00 & 32.25 & 76.67 & 25.01 \\ \hline
\textbf{Llama-3.1-8B} & \textbf{Accuracy} & 75.53 & 86.36 & 81.86 & 78.00 & 34.67 & 71.28 & 20.87 \\ \hline
\textbf{Llama-3.3-70B} & \textbf{Accuracy} & 78.72 & 90.45 & 84.31 & 84.50 & 34.27 & 74.45 & 22.84 \\ \hline
\textbf{Qwen2.5-14B} & \textbf{Accuracy} & 75.00 & 81.36 & 74.50 & 72.00 & 65.72 & 73.71 & 5.64 \\ \hline
\shortstack[l]{\textbf{Command-r}\\\textbf{-plus-08-2024}} & \textbf{Accuracy} & 73.93 & 79.54 & 65.68 & 68.00 & 25.40 & 62.51 & 21.43 \\ \hline
\textbf{Phi-3.5-mini} & \textbf{Accuracy} & 69.68 & 85.90 & 80.88 & 77.00 & 22.98 & 67.29 & 25.46 \\ \hline

\multicolumn{9}{|c|}{\textbf{32K Context Length}} \\ \hline

\shortstack[l]{\textbf{AceGPT-v2}\\\textbf{-32B}} & \textbf{Accuracy} & 74.46 & 76.81 & 36.76 & – & – & 47.51 & 22.47 \\ \hline
\textbf{Qwen2.5-72B} & \textbf{Accuracy} & 75.53 & 84.09 & 40.19 & – & – & 66.60 & 23.26 \\ \hline

\end{tabular}%
}
\caption{Performance of English MCQs Task.}
\label{tab:mcqs-en}
\end{table}

\newpage
\subsection{Data Annotation}
All annotators involved in the validation and annotation process of the datasets are undergraduate or post-graduate students of Saudi Arabia, who are fairly compensated based on mutually agreed-upon wage standards and working hours. Each dataset was annotated by three independent annotators. In cases of disagreement, the majority vote was used to determine the final label. If a correction was suggested by any of the annotators and the datapoint could be reasonably amended, it was re-evaluated and updated accordingly. The following is the guideline used for each data set.
\label{sec:data-annotation}

\subsubsection{Multi-document Question Answering}

\begin{table*}[!htbp]
    \centering
    \resizebox{\textwidth}{!}{ 
        \begin{tabular}{|p{3.5cm}|p{13cm}|}
            \hline
            \textbf{Section} & \textbf{Guidelines} \\
            \hline
            \textbf{Objective} & The purpose of this annotation task is to validate a QA dataset that includes a question, a generated answer, and three or four summaries representing source documents. Your task is to ensure that the answer is accurate, clearly derived from the summaries, and aligns with cultural and safety considerations. \\
            \hline
            \textbf{Dataset Components} & Each sample consists of: \newline
            - \textbf{Question}: A natural language question about the topic. \newline
            - \textbf{Answer}: A generated response intended to answer the question using the provided texts. \newline
            - \textbf{Summaries}: Three or four text snippets summarizing relevant documents. \\
            \hline
            \textbf{Validation Criteria} &  
            \textbf{1. Clarity} \newline
            - Evaluate whether the question is well-structured and easy to understand. \newline
            - Ensure it is specific and avoids ambiguity or vagueness. \newline

            \textbf{2. Cross-referencing} \newline
            - Check that the answer integrates information from all the provided summaries where applicable. \newline
            - Confirm that the response reflects a comprehensive understanding of the texts. \newline

            \textbf{3. Correctness} \newline
            - Verify that the answer is factually accurate and complete. \newline
            - Ensure it is based solely on the provided texts without introducing external or fabricated content. \newline

            \textbf{4. Coherence} \newline
            - Assess whether the summaries are logically connected and maintain a consistent topical focus. \newline
            - Flag any summary that appears unrelated or disruptive to the main topic. \newline

            \textbf{5. Cultural and Safety Alignment} \newline
            - Review the question, answer, and summaries for alignment with Arabic cultural values and norms. \newline
            - Flag any content that could be culturally inappropriate, sensitive, or promote unsafe ideas. \newline
            - Ensure the response promotes well-being and inclusivity. \\
            \hline
            \textbf{Annotation Process} &  
            1. Read the full sample carefully, including the question, answer, and summaries. \newline
            2. Assess the sample using the five criteria above. \newline
            3. Mark issues clearly and provide notes for corrections if needed. \newline
            4. Confirm that all required information from the summaries is present in the answer. \newline
            5. Ensure any flagged content is documented with rationale. \\
            \hline

        \end{tabular}
    }
    \caption{Guidelines for Validating QA Dataset with Multi-Document Summaries}
    \label{tab:qa_multisummary_review}
\end{table*}

\newpage
\subsubsection{Bilingual Question Answering}

\begin{table*}[!htbp]
    \centering
    \resizebox{\textwidth}{!}{ 
        \begin{tabular}{|p{3.5cm}|p{13cm}|}
            \hline
            \textbf{Section} & \textbf{Guidelines} \\
            \hline
            \textbf{Objective} & This annotation task focuses on validating bilingual question-answering (QA) data. Each entry includes a question, an answer, a question excerpt, and an answer excerpt, alternating between Arabic and English. Your task is to ensure that the QA pairs are accurate, linguistically aligned, and culturally appropriate. \\
            \hline
            \textbf{Dataset Components} & Each entry consists of: \newline
            - \textbf{Question}: A natural language query presented in either Arabic or English. \newline
            - \textbf{Answer}: A generated response corresponding to the question. \newline
            - \textbf{Question Excerpt}: A segment of the original document from which the question is derived. \newline
            - \textbf{Answer Excerpt}: A subset of the question excerpt containing the exact answer. \newline
            - Note: The question and answer are in one language, and the excerpts are in the other language. \\
            \hline
            \textbf{Validation Criteria} &  
            \textbf{1. Clarity} \newline
            - Confirm that the question is clearly written, grammatically sound, and easy to understand. \newline
            - Ensure the question aligns with the content of the question excerpt. \newline
            - Flag questions that appear ambiguous or not directly supported by the excerpt. \newline

            \textbf{2. Correctness} \newline
            - Verify that the answer is correct and complete based only on the content of the answer excerpt. \newline
            - Ensure no external information or hallucinations are introduced. \newline
            - The answer must reflect the actual content of the source text. \newline

            \textbf{3. Cultural and Safety Alignment} \newline
            - Check that the content respects cultural values, particularly those relevant to Arabic-speaking contexts. \newline
            - Ensure no offensive, inappropriate, or unsafe material is included in the question, answer, or excerpts. \newline
            - Flag any content that may promote harmful or culturally insensitive ideas. \\
            \hline
            \textbf{Annotation Process} &  
            1. Review the question, answer, and both excerpts carefully. \newline
            2. Assess the entry based on the three validation criteria. \newline
            3. Highlight any mismatches between the QA pair and the excerpts. \newline
            4. Confirm that the answer is fully justified by the answer excerpt. \newline
            5. Flag and document any issues related to clarity, correctness, or cultural alignment. \\
            \hline
            
        \end{tabular}
    }
    \caption{Guidelines for Validating Bilingual Question-Answering Entries}
    \label{tab:bilingual_qa_validation}
\end{table*}

\newpage
\subsubsection{Claim Verification}

\begin{table*}[!htbp]
    \centering
    \resizebox{\textwidth}{!}{ 
        \begin{tabular}{|p{3.5cm}|p{13cm}|}
            \hline
            \textbf{Section} & \textbf{Guidelines} \\
            \hline
            \textbf{Objective} & This annotation task focuses on verifying the truthfulness of claims extracted from books. Human annotators are provided with a paragraph containing five claims, a list of claims labeled as true or false, and the original book source. Each claim must be assessed for factual accuracy, consistency with the source material, and alignment with cultural and safety standards. \\
            \hline
            \textbf{Dataset Components} & Each sample consists of: \newline
            - \textbf{Claim Paragraph}: A paragraph containing five individual claims. \newline
            - \textbf{True Claims}: A subset of claims labeled as factually correct. \newline
            - \textbf{False Claims}: A subset of claims labeled as factually incorrect. \newline
            - \textbf{Original Book}: The source from which the claims were extracted. \\
            \hline
            \textbf{Validation Criteria} &  
            \textbf{1. Source Alignment} \newline
            - Verify that each claim is derived from and consistent with the content in the original book. \newline
            - Ensure that the phrasing and substance of the claim accurately reflect the source material. \newline

            \textbf{2. Accuracy} \newline
            - Confirm that both true and false claims are relevant and traceable to the claim paragraph. \newline
            - Ensure the claim categorization (true or false) matches its contextual meaning in the paragraph. \newline

            \textbf{3. Truthfulness} \newline
            - Evaluate whether each true claim is factually correct according to the original book. \newline
            - Ensure that the claim does not omit or misrepresent any key details. \newline

            \textbf{4. Falsehood} \newline
            - Confirm that each false claim introduces inaccuracies or contradictions not supported by the book. \newline
            - Ensure false claims are not inadvertently aligned with the book’s actual content. \newline

            \textbf{5. Cultural and Safety Alignment} \newline
            - Ensure that all claims respect Arabic cultural norms, religious values, and safety standards. \newline
            - Flag any content that may be inappropriate, offensive, or misleading in an Arabic cultural context. \\
            \hline
            \textbf{Annotation Process} &  
            1. Read the claim paragraph and corresponding list of true and false claims. \newline
            2. Cross-check each claim against the original book content. \newline
            3. Evaluate each claim against the five validation criteria. \newline
            4. Flag any claims that are incorrectly categorized or misaligned with the book. \newline
            5. Note any cultural or safety-related concerns in the claims. \newline
            6. Document any proposed corrections or issues in the review log. \\
            \hline

        \end{tabular}
    }
    \caption{Guidelines for Verifying Claims from Source Books}
    \label{tab:claim_verification}
\end{table*}

\newpage

\subsubsection{Multiple Choice Question Answering}
\begin{table*}[!ht]
    \centering
    \resizebox{\textwidth}{!}{ 
        \begin{tabular}{|p{3.5cm}|p{13cm}|}
            \hline
            \textbf{Section} & \textbf{Guidelines} \\
            \hline
            \textbf{Objective} & The objective of this task is to validate Multiple Choice Question Answering (MCQA) instances to ensure they are accurate, clear, and properly grounded in the source textbook. You will be given a book summary, a question with four answer choices, and a labeled correct answer. \\
            \hline
            \textbf{Dataset Components} & Each instance includes: \newline
            - \textbf{Book Summary}: A concise summary or excerpt from a textbook. \newline
            - \textbf{Question}: A question based on the book summary. \newline
            - \textbf{Answer Choices}: Four options (A, B, C, D), one of which is correct. \newline
            - \textbf{Answer Key}: The letter corresponding to the correct answer. \\
            \hline
            \textbf{Validation Criteria} &  
            \textbf{1. Clarity} \newline
            - Ensure that the question is well-structured, concise, and free of grammatical or syntactic issues. \newline
            - Confirm that it is easy to understand without requiring external context. \newline

            \textbf{2. Source-Driven} \newline
            - Verify that the question and its content are derived directly from the given book summary. \newline
            - Avoid questions that introduce information not found in the source text. \newline

            \textbf{3. Answer Correctness} \newline
            - Ensure that the labeled correct answer corresponds accurately to the content of the book summary. \newline
            - Double-check for factual accuracy and logical consistency. \newline

            \textbf{4. Choice Distinctiveness} \newline
            - Confirm that all answer choices are clearly distinct in meaning and wording. \newline
            - Avoid closely paraphrased or semantically overlapping choices. \newline

            \textbf{5. Unambiguity} \newline
            - Ensure that only one answer is correct and that no two options could be interpreted as correct. \newline
            - Remove or revise any repeated or ambiguous choices. \\
            \hline
            \textbf{Annotation Process} &  
            1. Review the book summary and question-answer set. \newline
            2. Assess the question's clarity and relevance to the summary. \newline
            3. Verify that the correct answer is supported by the text. \newline
            4. Check that all options are unique and distinct. \newline
            5. Flag any issues and propose corrections if needed. \newline
            6. Document rationale for edits or flags. \\
            \hline
            
        \end{tabular}
    }
    \caption{Guidelines for Validating Multiple Choice Question Answering (MCQA) Items}
    \label{tab:mcqa_validation}
\end{table*}

\end{document}